\documentclass{article}

\usepackage{microtype}
\usepackage{graphicx}
\usepackage{subfigure}
\usepackage{booktabs} 

\usepackage{hyperref}
\usepackage{paralist}

\usepackage[noend]{algorithmic}

\usepackage[accepted]{icml2024}

\usepackage{amsmath}
\usepackage{amssymb}
\usepackage{mathtools}
\usepackage{amsthm}

\usepackage[capitalize,noabbrev]{cleveref}

\theoremstyle{plain}
\newtheorem{theorem}{Theorem}[section]

\theoremstyle{definition}

\newtheorem{assumption}[theorem]{Assumption}
\theoremstyle{remark}

\usepackage[textsize=tiny]{todonotes}

\usepackage{caption}
\usepackage{adjustbox}
\usepackage{multirow}
\usepackage{bm}
\usepackage{makecell}
\usepackage{float}
\usepackage{colortbl}
\definecolor{codegreen}{rgb}{0,0.6,0}
\definecolor{codegray}{rgb}{0.5,0.5,0.5}
\definecolor{codepurple}{rgb}{0.58,0,0.82}
\definecolor{backcolour}{rgb}{0.95,0.95,0.92}
\definecolor{bestcolor}{RGB}{252, 229, 205}
\newcommand{\best}[1]{\cellcolor{bestcolor}{\textbf{#1}}}
\definecolor{secondcolor}{RGB}{243, 243, 243}
\newcommand{\second}[1]{\cellcolor{secondcolor}{\textbf{#1}}}
\definecolor{darkergreen}{RGB}{21, 152, 56}
\definecolor{red2}{RGB}{252, 54, 65}
\usepackage{pifont}
\newcommand{\cmark}{\textcolor{darkergreen}{\ding{51}}}
\newcommand{\xmark}{\textcolor{red2}{\ding{55}}}

\icmltitlerunning{Listwise Reward Estimation for Offline Preference-based Reinforcement Learning}

\begin{document}

\twocolumn[
\icmltitle{Listwise Reward Estimation for \\ Offline Preference-based Reinforcement Learning}



\icmlsetsymbol{equal}{*}

\begin{icmlauthorlist}
\icmlauthor{Heewoong Choi}{yyy}
\icmlauthor{Sangwon Jung}{yyy}
\icmlauthor{Hongjoon Ahn}{yyy}
\icmlauthor{Taesup Moon}{yyy,zzz}
\end{icmlauthorlist}

\icmlaffiliation{yyy}{Department of Electrical and Computer Engineering, Seoul National University}
\icmlaffiliation{zzz}{ASRI/INMC/IPAI/AIIS, Seoul National University}


\icmlcorrespondingauthor{Taesup Moon}{tsmoon@snu.ac.kr}

\icmlkeywords{Machine Learning, ICML}

\vskip 0.3in
]



\printAffiliationsAndNotice{}  

\begin{abstract}
In Reinforcement Learning (RL), designing precise reward functions remains to be a challenge, particularly when aligning with human intent. Preference-based RL (PbRL) was introduced to address this problem by learning reward models from human feedback. However, existing PbRL methods have limitations as they often overlook the \textit{second-order} preference that indicates the relative strength of preference. In this paper, we propose Listwise Reward Estimation (LiRE), a novel approach for offline PbRL that leverages second-order preference information by constructing a Ranked List of Trajectories (RLT), which can be efficiently built by using the same ternary feedback type as traditional methods. To validate the effectiveness of LiRE, we propose a new offline PbRL dataset that objectively reflects the effect of the estimated rewards. Our extensive experiments on the dataset demonstrate the superiority of LiRE, \textit{i.e., }outperforming state-of-the-art baselines even with modest feedback budgets and enjoying robustness with respect to the number of feedbacks and feedback noise. Our code is available at \href{https://github.com/chwoong/LiRE}{https://github.com/chwoong/LiRE}
\end{abstract}

\section{Introduction}
\label{intro}
Reinforcement Learning (RL) has demonstrated considerable success in various domains such as robotics \cite{haarnoja2018soft, kalashnikov2018scalable}, game \cite{silver2017mastering, mnih2013playing, vinyals2019grandmaster}, autonomous driving \cite{kiran2021deep}, and real-world tasks \cite{tan2018sim, chebotar2019closing}.
An essential component of RL is to define suitable and precise reward functions so that an RL agent can be trained successfully  \cite{wirth2017survey}.
However, designing the reward function is time-consuming, especially if we want to align it with human intent \cite{hejna2023inverse}.

This shortcoming has led to research on learning the reward model from human feedback without explicitly designing the reward function.
While expert demonstration is one type of human feedback \cite{abbeel2004apprenticeship}, recent papers use preference feedback on which of a pair of trajectory segments is preferred since it is a significantly easier type of feedback to collect \cite{kaufmann2023survey, casper2023open}.
More specifically, the common approach for the Preference-based RL (PbRL) consists of two steps: (1) learn a reward model using preference feedback from trajectory segment pairs, then (2) apply ordinary RL algorithms with the learned reward model.
After successfully training a robot agent with PbRL \cite{christiano2017deep}, it was shown that novel behaviors aligned with human intent, \textit{e.g.,} backflips, can also be learned \cite{lee2021pebble}, while learning such behavior would be extremely hard from explicitly hand-coded rewards.
The PbRL framework has gained popularity in both online \cite{park2021surf, liang2021reward} and offline \cite{kim2022preference, shin2022benchmarks, an2023direct, hejna2023inverse} settings, in which the former allows the agents to interact with their environments, while the latter does not.

\begin{figure*}[t]
  \centering
  \includegraphics[width=1.0\textwidth]{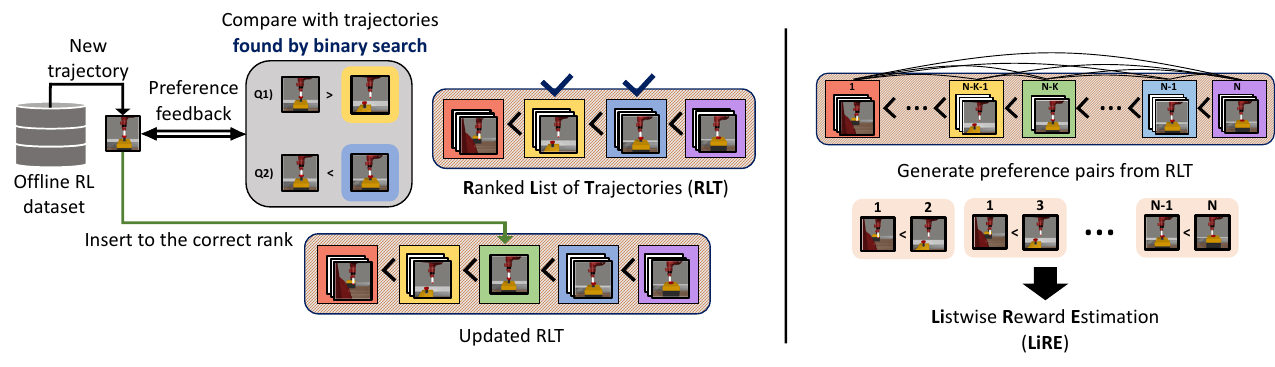}
  \captionsetup{skip=2mm}
  \caption{\textbf{An overview of LiRE.} The figure shows an example of a \textit{button-press-topdown} task. We sample a trajectory segment and sequentially obtain the preference feedback for existing trajectories in RLT. We use binary search to find the correct rank (left) efficiently. Multiple preference pairs are generated from RLT to learn the reward model (right).}
  \label{fig:example}
  \vspace{-.1in}
\end{figure*}

In this paper, we focus on the offline PbRL setting, in which the goal is to find an optimal policy solely from the \textit{previously collected} preference feedbacks on the pairs of trajectories obtained from some past, fixed policy.
This setting is challenging since the preference feedback cannot be actively collected on the trajectories generated by the current, updated policy.
Hence, developing effective methods for collecting maximally informative preference feedback data from the past policy as well as devising efficient reward learning schemes is indispensable.

The current norm is to collect ternary preference feedback (\textit{i.e.,} more/less/equally preferred) for \textit{independently} sampled pairs of trajectories, and then employ the standard Bradley-Terry (BT) model \cite{bradley1952rank} on the collected data to learn the reward function. 
While the above approach was shown to be effective to some extent, a critical limitation also exists. Namely, due to the independent sampling of the pairs of trajectories and simple ternary feedback, the \textit{second-order} preference, which stands for the \textit{relative} strengths of the preferences, cannot be utilized.
There exists a long line of work in several areas asserting that utilizing such second-order preference is indeed effective for more accurate learning \cite{xia2008listwise, touvron2023llama, hwang2023sequential, song2023preference}. 
However, the majority of these works presume the availability of more sophisticated preference feedback types, which are considerably more laborious and expensive to obtain than the above-mentioned ternary feedback.

To that end, we propose to construct a Ranked List of Trajectories (RLT) while collecting preference feedback data to exploit the second-order preference when learning a reward function.
The key novelty and strength of our method is to use \textit{exactly the same} feedback type and budget as before and not require any additional sophistication in collecting the preference feedback.
As outlined in Figure \ref{fig:example}, the main idea of building such a ranked list is to sample \textit{a} trajectory and \textit{sequentially} obtain the preference feedback by comparing it with existing trajectories in the ranked list multiple times to find its correct rank in the list.
Hence, our method ends up sampling \textit{fewer} trajectories for a fixed feedback budget compared to the conventional independent pair sampling.
However, once the complete RLT is built, the second-order preference can be extracted and exploited for estimating the reward function, which, as we show in our experiments, results in a significant performance boost of offline PbRL.

The superiority of our method, dubbed as \textbf{LiRE} (\textbf{Li}stwise \textbf{R}eward \textbf{E}stimation), is demonstrated through extensive experimental validation.
We first created an offline RL dataset using Meta-World \citep{yu2020meta} and DeepMind Control Suite (DMControl) \cite{tassa2018deepmind} environments that can objectively compare the reward estimation quality of offline PbRL methods.
This is motivated by \cite{li2023survival}, which pointed out that the offline RL performance can be high in some popular benchmark datasets even with \textit{wrong} or \textit{constant} reward functions.
On our proposed datasets, we show that many tasks \textit{cannot} be properly learned with existing offline PbRL methods, even with a large preference feedback budget.
In contrast, we showcase our LiRE can outperform those baselines on most of the tasks with significant margins even with a modest preference feedback budget.
We conduct comprehensive experimental analyses to investigate the impact of several factors,
including the score function of the BT model, the number of preference feedbacks, and the number of trajectories in the RLT.
The experimental results show that the degree to which second-order information is utilized has a significant positive impact on the performance of offline PbRL.
Furthermore, the results of the real human preference feedback experiments, along with experiments on the level of preference feedback noise and feedback granularity, demonstrate the effectiveness of LiRE in practical scenarios.
These analyses provide substantial evidence supporting the strength and robustness of LiRE.
\section{Related Works}


\subsection{Offline Preference-based RL}

Due to the difficulty of defining rewards in reinforcement learning \cite{sutton2018reinforcement, mckinney2023fragility}, PbRL uses comparison information between trajectories to learn a reward function \cite{christiano2017deep, furnkranz2012preference, wilson2012bayesian, akrour2012april, ouyang2022training, stiennon2020learning}.
However, the human preference feedback required for PbRL is expensive to obtain.
Thus, several PbRL approaches have been devised to reduce the number of expensive human feedbacks, such as using additional expert demonstrations \cite{ibarz2018reward}, meta-learning \cite{hejna2023few}, semi-supervised learning or data augmentation \cite{park2021surf}, unsupervised pre-training \cite{lee2021pebble}, exploration based on reward uncertainty \cite{liang2021reward}, and using sequential preference ranking \cite{hwang2023sequential}.
Offline PbRL assumes a more challenging problem setting where agents cannot interact with the environment, unlike online PbRL where preference feedback can be obtained while interacting with the environment.

In offline PbRL, the two kinds of data are provided: offline data obtained from an unknown policy and preference feedbacks on the pairs of trajectories.
Also, traditional offline PbRL methods have two phases;
they train a reward model using the preference feedback and then perform RL with the trained reward model without interacting with the environment.
On the other hand, recent works propose performing offline PbRL without the reward model by directly optimizing policies \cite{an2023direct, kang2023beyond}, or learning state-action value function or regret from preference labels \cite{hejna2023inverse, hejna2023contrastive}.
However, due to the constraint of no interaction with the environment, obtaining the most informative preference feedback from the offline dataset is as important as developing a new training method without the reward model or designing the structure of the reward model well (\textit{e.g.,} non-Markovian reward modeling \cite{kim2022preference}).
An active query selection method has been proposed to obtain informative preference pairs \cite{shin2022benchmarks}, but their method did not attempt to obtain second-order preference.

Most offline PbRL papers have validated their algorithms on the D4RL dataset \cite{fu2020d4rl}.
However, it has been shown that typical offline RL algorithms can produce good policies on D4RL even with a completely wrong reward (\textit{e.g.,} zero, random, negative reward) due to the pessimism and survival instinct of offline RL algorithms \cite{shin2022benchmarks, li2023survival}.
Hence, to properly evaluate how well offline PbRL algorithms learn the reward model, we need to validate them on a new dataset, on which the policy cannot be easily learned due to survival instincts.

\subsection{Second-order Preference Feedback}
While typical approaches in PbRL only focus on the first-order preference (\textit{i.e.,} ternary labels including bad, equal, and good), several approaches in the NLP and RL domains have recently been proposed to utilize second-order preference about the relative difference between preferences.
One approach is to directly obtain a relative rating for each trajectory pair (\textit{e.g.,} significantly better or slightly better) or an absolute rating for each trajectory (\textit{e.g.,} very good or good) \cite{touvron2023llama, cao2021weak, white2023rating}. However, the more granular the preferences are, the more expensive they are than just ternary labels.

There is a rich Learning-to-Rank literature that learns the ranking given second-order preference feedback in the form of absolute ratings \cite{burges2005learning, xia2008listwise, xu2007adarank, swezey2021pirank}, but they do not address how to obtain second-order preference only with ternary labels.
Another approach is to obtain the second-order preference between samples from a fully-ranked list for multiple trajectories \cite{chen2022human, palan2019learning, zhu2023principled, song2023preference, myers2022learning, biyik2019green, brown2019extrapolating}.
However, they do not address how to efficiently obtain the fully-ranked list in terms of the number of feedbacks.
Since naively constructing a fully-ranked list would require a large number of feedbacks that increase quadratically with the number of trajectories, developing a more efficient list construction method is crucial.

Accordingly, some recent studies have developed how to obtain partially-ranked lists that only know the rankings among a few trajectories \cite{zhao2023slic, hwang2023sequential}.
Perhaps, one of the closest research to ours is Sequential Preference Ranking (SeqRank) \cite{hwang2023sequential}, which sequentially collects the preference feedback between a newly observed segment and a previously collected segment.
However, since their method builds partially-ranked lists rather than fully-ranked lists, the short length of the lists limits the ability to fully utilize second-order information.

\section{Preliminaries}\label{sec:prelim}
An RL algorithm considers a Markov decision process (MDP) and aims to find the optimum policy that maximizes the cumulative discounted rewards.
MDP is defined by a tuple $(S, A, P, r, \gamma)$ where $S, A$ are state, action space, $P=P(\cdot|s,a)$ is the environment transition dynamics, $r=r(s,a)$ is reward function, and $\gamma$ is discount factor.
In offline PbRL, we assume that we do not know the true reward $r$, but we have a pre-collected dataset that is a set of tuples, $D_o \coloneqq \{(s,a,s^\prime)|(s, a) \sim \mu, s^\prime \sim P(\cdot|s, a)\}$.
In general, the policy $\mu$  from which the data was collected is unknown.
We are allowed to ask for preference feedbacks to obtain preference labels for two distinct trajectory segments sampled from $D_s \coloneqq \{\sigma \,|\, \sigma=(s_{0}, a_{0}, s_{1}, a_{1}, \cdots,s_{T-1},a_{T-1}), (s_{t},a_{t},s_{t+1}) \in D_o\}$.
Annotators assign a ternary label $l$ given a pair of segments $\sigma_1, \sigma_2 \in D_s$;
$l=0$ indicates that $\sigma_{1}$ is preferred over $\sigma_{2}$ (\textit{i.e.,} $\sigma_1 \succ \sigma_2$), $l=1$ indicates the opposite preference (\textit{i.e.,} $\sigma_1 \prec \sigma_2$), and $l=0.5$ indicates that $\sigma_{1}$ and $\sigma_{2}$ are equally preferred (\textit{i.e.,} $\sigma_1 = \sigma_2$).

The goal of acquiring preference labels is to learn the unknown reward function.
Conventional offline PbRL methods use a preference model that defines the probability that one segment is better than the other as
\begin{equation}
    \label{eq:pl model}
    P_\theta(\sigma_1 \succ \sigma_2) = \frac{\phi\bigl(r_\theta(\sigma_1)\bigr)}{\phi\bigl(r_\theta(\sigma_1)\bigr)+\phi\bigl(r_\theta(\sigma_2)\bigr)}
\end{equation}
in which $r_\theta(\sigma_i)=\sum_{(s_t,a_t) \in \sigma_i}r_\theta(s_t,a_t)$ and $\theta$ is the parameter of the reward model.
The score function $\phi(x)=\text{exp}(x)$ is commonly used in the BT model \cite{bradley1952rank}.
Given the trajectory segment preference dataset, $D_{\textit{pref}} \coloneqq \{(\sigma_{i_1}, \sigma_{i_2}, l_i)\}_{i=1}^K$, the parameter $\theta$ is learned by minimizing following cross-entropy loss:
\begin{multline}
    \label{eq:BT model}
    L(\theta) = -\mathop{\mathbb{E}}_{\substack{(\sigma_{i_1}, \sigma_{i_2}, l_i) \\\in D_{\textit{pref}}}}\,
    \Bigr[(1-l_i)\,\log P_\theta\left(\sigma_{i_1} \succ \sigma_{i_2}\right) + \\
    l_i\,\log P_\theta\left(\sigma_{i_1} \prec \sigma_{i_2}\right)\Bigr].
\end{multline}

\section{LiRE: Listwise Reward Estimation}
As mentioned in \cref{intro}, the conventional offline PbRL approaches cannot utilize the second-order information of the preference feedback. In order to describe our method, we begin by stating the mild assumptions we make.
\begin{assumption}
\label{assumption}
\textbf{(Completeness)}
For any two segments $\sigma_i, \sigma_j$,
the human feedbacks are provided in the following three ways, $\sigma_i \succ \sigma_j$ or $\sigma_i \prec \sigma_j$ or $\sigma_i = \sigma_j$.\\
\textbf{(Transitivity)}
For any three segments $\sigma_i, \sigma_j$, and $\sigma_k$,
if $\sigma_i = \sigma_j$ and $\sigma_j = \sigma_k$, then $\sigma_i = \sigma_k$.
Also, if $\sigma_i \succ \sigma_j$ and $\sigma_j \succ \sigma_k$, then $\sigma_i \succ \sigma_k$.\\
\noindent \textit{Remarks: }
These assumptions are a generalization of SeqRank \cite{hwang2023sequential} to include equal labels.
While the transitivity assumption may not always hold in practice, we demonstrate that our method is robust both in the presence of feedback noise (\cref{exp:noise}) and in real human experiments (\cref{exp:human}), even when the transitivity assumption may not hold.
\end{assumption}

\subsection{Constructing a Ranked List of Trajectories (RLT)}

Our goal is to obtain an RLT in which the segments $\sigma$ are ordered by their level of preference.
We represent RLT, $L$, in the following form:
$$
L = [g_1 \prec g_2 \prec \cdots \prec g_s],
$$
in which $g_i = \{\sigma_{i_1}, \cdots, \sigma_{i_k}\}$ is a \textit{group} of segments with the same preference level and $s$ is the number of groups in the list.
Namely, if $m > n$, we note any segment $\sigma_i \in g_m$ is preferred over any segment $\sigma_j \in g_n$.

Since we assume to have exactly the same type of ternary feedback defined in Section \ref{sec:prelim}, we cannot build RLT by obtaining the listwise feedback at once. Hence, we construct by sequentially obtaining the labels as we describe below.

We start with an initial list $[\{\sigma_1\}]$ by selecting a random segment $\sigma_1$ from $D_s$.
We then repeat the process of sequentially sampling the new segment $\sigma_2, \sigma_3, \cdots \in D_s$ and placing it in the appropriate position in the list until the feedback budget limit is reached.
To place a newly sampled $\sigma_i$ in the RLT, we compare it with a segment $\sigma_k \in g_m$ for some group $g_m$ in the list and obtain the ternary preference feedback. Depending on the feedback, we proceed as follows:
\begin{compactitem}
    \item If $\sigma_i = \sigma_k$, add $\sigma_i$ to the group $g_m$.
    \item If $\sigma_i \prec \sigma_k$, find the position within $g_{1}, \cdots, g_{m-1}$.
    \item If $\sigma_i \succ \sigma_k$, find the position within $g_{m+1}, \cdots, g_s$.
\end{compactitem}
For the latter two cases, we use a binary search so that we can recursively find the correct group for each segment. 
Namely, the RLT construction algorithm is based on a \textit{binary insertion sort} and the pseudocode is summarized in \cref{alg:pref-list} (Appendix).
We note that while we can also adopt merge sort or quick sort to construct an RLT after collecting multiple segments, if we already have a partially constructed RLT, binary insertion sort would be more feedback-efficient.

\begin{table}[!t]
\caption{Feedback efficiency and sample diversity of independent pairwise sampling, SeqRank, and RLT. 
}
\label{tab: efficiency}
\resizebox{\columnwidth}{!}{
\begin{tabular}{l|ccc}
                    & Independent & SeqRank & RLT \\ \hline
Feedback efficiency & 1        & 1.392   & $\mathcal{O}(M/\log{M})$    \\
Sample diversity   & 2        & $\mathcal{O}(1)$    & $\mathcal{O}(1/\log{M})$
\end{tabular}
}
\vspace{-.2in}
\end{table}

\textbf{Feedback efficiency and sample diversity}
Note that by design, we need to obtain \textit{multiple} preference feedbacks for each new segment $\sigma_i$.
Therefore, for a fixed feedback budget, our method samples fewer segments. However, from the constructed RLT, we can generate \textit{many} preference pairs by exploiting the second-order information encoded in the list; namely, $\sigma_i$ is preferred to \textit{all} the segments in the groups that rank lower than the group that $\sigma_i$ belongs to.

To that end, we analyze the feedback efficiency and sample diversity of RLT.
\textbf{Feedback efficiency} is defined in SeqRank \cite{hwang2023sequential} as the ratio of the number of total preference pairs generated to the number of preference feedbacks obtained.
We also define \textbf{sample diversity} as the ratio of the total number of sampled segments to the number of preference feedbacks obtained.
Suppose we obtain preference feedbacks until we collect a total of $M$ segments in the preference dataset.
Constructing an RLT with $M$ segments requires $\mathcal{O}(M\log M)$ feedbacks because we use an efficient sorting method based on binary search.
In this case, the number of all possible preference pairs (including ties) that can be generated from the RLT is $\binom{M}{2}$.
\cref{tab: efficiency} summarizes the feedback efficiency and sample diversity of independent pairwise sampling, SeqRank, and RLT.
Note our method has a faster rate of increase in the feedback efficiency even with diminishing sample diversity as the number of segments $M$ in RLT increases.

\textbf{Constructing multiple RLTs}
\label{Q_budget}
\cref{alg:pref-list} places all the segments in a single ranked list.
Instead of constructing one long list, we devise a variant that generates multiple lists by setting a limit ($Q$) on the feedback budget for each list.
The reason for generating multiple lists is that as the length of the list increases, the number of preference feedbacks required by the binary search process increases.
Hence, we increase the sample diversity within the total feedback budget by generating multiple RLTs.

\subsection{Listwise Reward Estimation from RLT}
\label{method:LiRE}
Once the RLT is constructed, we construct the preference dataset, $D_{l}=\{(\sigma_{i_1},\sigma_{i_2},l_i)\}_{i=1}^{K}$ with all the pairs we can obtain from the RLT.
Specifically, when $\sigma_{i_1} \in g_m$ and $\sigma_{i_2} \in g_n$, the preference label $l_i$ is as follows: $l_i=0.5$ if $m=n$, $l_i=0$ if $m>n$, and $l_i=1$ if $m<n$.
The key difference from traditional pairwise PbRL methods is that, instead of independently sampling segment pairs, we derive preference pairs from the RLT.
To compare with the independent sampling, suppose that the RLT has segments with the relationship, $\sigma_a < \sigma_b < \sigma_c$.
If we sample all pairs from the RLT, then $(\sigma_a, \sigma_b, 1), (\sigma_b, \sigma_c, 1), (\sigma_a, \sigma_c, 1) \in D_{l}$.
From these preference pairs, it can be inferred that the degree to which $\sigma_c$ is preferred over $\sigma_a$ is stronger than the degree to which $\sigma_c$ is preferred over $\sigma_b$.
Consequently, the reward model trained with pairwise loss in \eqref{eq:BT model} can learn second-order preference between each pair of segments.
In contrast, the reward model learned from independent sampling cannot learn second-order preference because each segment is not compared to multiple other segments.

We use pairwise loss in our main experiments, but we can also train the reward model with listwise loss since the segments are ranked in the RLT.
To train the reward model with listwise loss, we assume the segments follow a Plackett-Luce model \cite{plackett1975analysis} which defines the probability distribution of objects in a ranked list.
We discuss listwise loss more in detail in \cref{appendix:listwise} --- but, our experimental results show that training with pairwise loss performs better than listwise loss in most cases.

Our proposed LiRE trains the reward model with linear score function $\phi(x)=x$ in \eqref{eq:pl model}.
The choice of linear score function has the same effect as setting the reward to be the exponent of the optimal reward value obtained through training with an exponential score function $\phi(x)=\text{exp}(x)$.
Therefore, the linear score function amplifies the difference in reward values, particularly in regions with high reward values, compared to the exponential score function.

\textbf{Bounding reward model}
If $\phi(x)=\text{exp}(x)$, then \textit{adding} a constant to the reward function $\hat{r}_\theta$ does not affect the resulting probability distribution.
To align the scaling of the learned $\hat{r}_\theta$ in ensemble reward models, a common choice for the reward model is using the Tanh activation, \textit{i.e., } $\hat{r}_\theta(\sigma) = \sum_{t}\hat{r}_\theta(s_t,a_t) = \sum_{t}\tanh(f_\theta(s_t, a_t))$ \cite{lee2021pebble, hejna2023inverse}, to bound the output of the reward model.

In the case of $\phi(x)=x$, \textit{scaling} the reward function by a constant does not affect the probability distribution.
Similarly, we use the same Tanh activation function for $\phi(x)=x$ to bound the output of the reward model.
Specifically, we set $\hat{r}_\theta(\sigma) = \sum_{t}\bigl(1+\tanh(f_\theta(s_t, a_t))\bigr) > 0$ to ensure that the probability defined in \eqref{eq:pl model} is positive.
\section{Experimental Results}
\subsection{Settings}
\textbf{Dataset}
Previous offline PbRL papers are evaluated mainly on D4RL \cite{fu2020d4rl}, but D4RL has the problem that RL performance can be high even when wrong rewards are used \cite{li2023survival, shin2022benchmarks}. To that end, 
we newly collect the offline PbRL dataset with Meta-World \cite{yu2020meta} and DeepMind Control Suite (DMControl) \cite{tassa2018deepmind} following the protocols of previous work: \texttt{medium-replay} dataset, \textit{e.g.,} \cite{yu2021conservative, mazoure2023contrastive, gulcehre2020rl} and \texttt{medium-expert} dataset, \textit{e.g.,} \cite{yu2021combo, sinha2022s4rl, hejna2023inverse, li2023survival}.

\begin{table*}[ht]
	\centering
	\small
	\caption{Average success rates on \texttt{medium-replay} dataset over six random seeds. We use $500$ and $1000$ preference feedbacks and report the average performance of the last five trained policies. The yellow and gray shading represent the best and second-best performances, respectively.}
	\label{tab:baselines}
	\begin{adjustbox}{max width=\textwidth}
		\begin{tabular}{c|l|rrrrrrrr}
			\toprule
			\multirow{2}{*}{\textbf{\makecell[c]{\# of\\ feedbacks}}} & \multirow{2}{*}{\textbf{Algorithm}} & \multirow{2}{*}{\textbf{\makecell[r]{button-press \\ -topdown}}} & \multirow{2}{*}{\textbf{box-close}} & \multirow{2}{*}{\textbf{dial-turn}} & \multirow{2}{*}{\textbf{sweep}} & \multirow{2}{*}{\textbf{\makecell[r]{button-press \\ -topdown-wall}}} & \multirow{2}{*}{\textbf{sweep-into}} & \multirow{2}{*}{\textbf{drawer-open}} & \multirow{2}{*}{\textbf{lever-pull}}  \\
			& & & & & & & \\
			\midrule
			\midrule
-  & IQL with GT rewards & 88.33 {\tiny $\pm$ 4.76}    & 93.40 {\tiny $\pm$ 3.10} & 75.40 {\tiny $\pm$ 5.47} & 98.33 {\tiny $\pm$ 1.87} & 56.27 {\tiny $\pm$ 6.32}         & 78.80 {\tiny $\pm$ 7.96} & 100.00 {\tiny $\pm$ 0.00}& 98.47 {\tiny $\pm$ 1.77} \\ \cmidrule{1-10}
\multirow{7}{*}{500} & MR          & 9.60 {\tiny $\pm$ 5.74}     & 10.33 {\tiny $\pm$ 8.23} & 50.20 {\tiny $\pm$ 8.51} & 79.80 {\tiny $\pm$ 13.36} & 0.13 {\tiny $\pm$ 0.50}          & 24.80 {\tiny $\pm$ 5.28} & 98.07 {\tiny $\pm$ 3.20} & 50.53 {\tiny $\pm$ 8.55} \\
& PT \cite{kim2022preference}         & 22.87 {\tiny $\pm$ 9.06}    & 0.33 {\tiny $\pm$ 1.16}  & \second{68.67} {\tiny $\pm$ 12.39} & 43.07 {\tiny $\pm$ 24.57} & 0.87 {\tiny $\pm$ 1.43}          & 20.53 {\tiny $\pm$ 8.26} & 88.73 {\tiny $\pm$ 11.64} & 82.40 {\tiny $\pm$ 22.69} \\
& OPRL \cite{shin2022benchmarks}       & 12.13 {\tiny $\pm$ 5.75}    & 4.73 {\tiny $\pm$ 3.24}  & 54.33 {\tiny $\pm$ 11.47} & \best{94.13} {\tiny $\pm$ 5.95} & 0.20 {\tiny $\pm$ 0.60}          & 25.87 {\tiny $\pm$ 8.58} & 94.13 {\tiny $\pm$ 6.41} & 54.67 {\tiny $\pm$ 12.79} \\
& DPPO \cite{an2023direct}       & 3.93 {\tiny $\pm$ 4.34}     & 10.20 {\tiny $\pm$ 11.47} & 26.67 {\tiny $\pm$ 22.23} & 10.47 {\tiny $\pm$ 15.84} & 0.80 {\tiny $\pm$ 1.51}          & 23.07 {\tiny $\pm$ 7.02} & 35.93 {\tiny $\pm$ 11.18} & 10.13 {\tiny $\pm$ 12.19} \\
& IPL \cite{hejna2023inverse}        & \second{34.73} {\tiny $\pm$ 13.92}    & 5.93 {\tiny $\pm$ 5.81}  & 31.53 {\tiny $\pm$ 12.50} & 27.20 {\tiny $\pm$ 23.81} & \second{8.93} {\tiny $\pm$ 9.84}          & \second{32.20} {\tiny $\pm$ 7.35} & 19.00 {\tiny $\pm$ 13.63} & 31.20 {\tiny $\pm$ 15.76} \\ 
& SeqRank \cite{hwang2023sequential}        & 17.6 {\tiny $\pm$ 11.94}    & \second{13.2} {\tiny $\pm$ 12.72}   & 65.6 {\tiny $\pm$ 12.84} & \second{83.4} {\tiny $\pm$  9.76} & 1.73 {\tiny $\pm$ 1.98}         & 25.67 {\tiny $\pm$ 11.02}  & \best{99.53}  {\tiny $\pm$ 0.36}&  \best{95.67} {\tiny $\pm$  4.04} \\ \cmidrule{2-10}
& LiRE (ours) & \best{67.20} {\tiny $\pm$ 18.97}    & \best{51.53} {\tiny $\pm$ 18.48} & \best{79.07} {\tiny $\pm$ 10.96} & 77.53 {\tiny $\pm$ 10.50} & \best{79.13} {\tiny $\pm$ 15.19}         & \best{49.13} {\tiny $\pm$ 15.85} & \second{99.40} {\tiny $\pm$ 1.65} & \best{95.67} {\tiny $\pm$ 6.26} \\
\midrule
\multirow{7}{*}{1000} & MR          & 9.27 {\tiny $\pm$ 5.30}& 17.07 {\tiny $\pm$ 9.56}& 59.07 {\tiny $\pm$ 7.57}& \best{90.80} {\tiny $\pm$ 9.74}& 0.60 {\tiny $\pm$ 1.87}& 26.07 {\tiny $\pm$ 8.57}& 96.47 {\tiny $\pm$ 4.02}& 50.87 {\tiny $\pm$ 10.89}  \\
& PT \cite{kim2022preference}         & 18.27 {\tiny $\pm$ 10.62} & 2.27 {\tiny $\pm$ 2.86}& 68.80 {\tiny $\pm$ 5.50}& 29.13 {\tiny $\pm$ 14.55} & 2.13 {\tiny $\pm$ 2.96}& 20.27 {\tiny $\pm$ 7.84}& 95.40 {\tiny $\pm$ 7.27}& 72.93 {\tiny $\pm$ 10.16} \\
& OPRL \cite{shin2022benchmarks}       & 11.00 {\tiny $\pm$ 7.84}& 15.07 {\tiny $\pm$ 11.19} & 51.33 {\tiny $\pm$ 10.08} & \second{85.53} {\tiny $\pm$ 5.43}& 0.33 {\tiny $\pm$ 0.75}& 28.27 {\tiny $\pm$ 6.40}& \second{99.20} {\tiny $\pm$ 1.42}& 53.20 {\tiny $\pm$ 6.67}\\
& DPPO \cite{an2023direct}       & 3.20 {\tiny $\pm$ 3.04}& 9.33 {\tiny $\pm$ 9.60}& 36.40 {\tiny $\pm$ 21.95} & 8.73 {\tiny $\pm$ 16.37} & 0.27 {\tiny $\pm$ 0.85}& 23.33 {\tiny $\pm$ 7.80}& 36.47 {\tiny $\pm$ 7.30}& 8.53 {\tiny $\pm$ 9.96} \\
& IPL  \cite{hejna2023inverse}       & \second{36.67} {\tiny $\pm$ 17.40} & 6.73 {\tiny $\pm$ 8.41}& 43.93 {\tiny $\pm$ 13.37} & 38.33 {\tiny $\pm$ 24.87} & \second{14.07} {\tiny $\pm$ 11.47} & \second{30.40} {\tiny $\pm$ 7.74}& 28.53 {\tiny $\pm$ 18.37} & 40.40 {\tiny $\pm$ 17.38} \\ 
& SeqRank \cite{hwang2023sequential}        & 13.93 {\tiny $\pm$ 8.11}   & \second{46.60}  {\tiny $\pm$12.53} & \second{70.67} {\tiny $\pm$ 7.58}& 74.93  {\tiny $\pm$22.35}&  2.47  {\tiny $\pm$2.67 }    &  29.33  {\tiny $\pm$ 11.59} &  98.6 {\tiny $\pm$ 3.92}&  \second{95.47}  {\tiny $\pm$ 3.86}\\ \cmidrule{2-10} 
& LiRE (ours) & \best{83.07} {\tiny $\pm$ 6.38}& \best{89.13} {\tiny $\pm$ 6.02}& \best{76.93} {\tiny $\pm$ 7.55}& 75.87 {\tiny $\pm$ 6.81}& \best{81.47} {\tiny $\pm$ 10.04} & \best{57.73} {\tiny $\pm$ 13.11} & \best{99.73} {\tiny $\pm$ 0.85}& \best{99.47} {\tiny $\pm$ 1.15} \\

		\bottomrule
		\end{tabular}
	\end{adjustbox}
\vspace{-.2in}
\end{table*}

The \texttt{medium-replay} dataset collects data from replay buffers used in online RL algorithms, such as the SAC \cite{haarnoja2018soft}, and the \texttt{medium-expert} dataset collects trajectories generated by the noisy perturbed expert policy.
We experiment on both datasets while our main analyses are done on \texttt{medium-replay}; see \cref{appendix-dataset} for complete details on constructing them.
The prior works \cite{shin2022benchmarks, anonymous2023efficient} have created datasets that consider survival instinct.
However, their dataset was evaluated with only $100$ or fewer preference feedbacks, whereas we use $500$, $1000$, or more feedbacks.

\textbf{Baselines}
In our experiments, we consider five baselines: Markovian Reward (MR), Preference Transformer (PT) \cite{kim2022preference}, Offline Preference-based Reward Learning (OPRL) \cite{shin2022benchmarks}, Inverse Preference Learning (IPL) \cite{hejna2023inverse}, and Direct Preference-based Policy Optimization (DPPO) \cite{an2023direct}.
MR refers to the method trained with the MLP layer with the Markovian reward assumption, which is the baseline model used in PT.
OPRL learns multiple reward models to select the query actively with the highest preference disagreement.
Lastly, IPL and DPPO are algorithms that learn policies without the reward model.

All the above five baselines belong to pairwise PbRL because they all train based on the BT model given first-order preference feedbacks sampled as independent pairs.
In addition to pairwise PbRL, we also compare with the sequential pairwise comparison method proposed by SeqRank \cite{hwang2023sequential}.

\textbf{Implementation details}
For LiRE, we use the linear score function and set $Q=100$ as the default feedback budget for each list.
Therefore, if the total number of feedbacks is $500$, then five RLTs will be constructed.
All baseline methods, including ours, can be applied to any offline RL algorithm, but, as in previous works, we use IQL \cite{kostrikov2021offline}.
The hyperparameters for each algorithm and the criteria for the equally preferred label threshold of scripted teacher can be found in the \cref{appendix-details}.

\begin{table}[!t]
    \small
    \caption{Average success rates on \texttt{medium-replay} with $500$ preference feedbacks. 
    }
    \label{tab:linear}
    \centering
    \tabcolsep=0.13cm
    \resizebox{\columnwidth}{!}{
    \begin{tabular}{cc|rrrr}
        \toprule
        RLT & $\phi(x)$ & \makecell{button-press\\-topdown} & box-close & dial-turn &  lever-pull \\ \midrule
        \xmark & $\text{exp}(x)$ & 9.60 {\tiny $\pm$ 5.74 }& 10.33 {\tiny $\pm$ 8.23 }& 50.20 {\tiny $\pm$ 8.51 }& 50.53 {\tiny $\pm$ 8.55 }\\
        \cmark & $\text{exp}(x)$ & 12.87 {\tiny $\pm$ 7.86 }& 22.73 {\tiny $\pm$ 10.40} & 65.87 {\tiny $\pm$ 9.46 }& 57.87 {\tiny $\pm$ 11.28}  \\
        \midrule
        \xmark & $x$ & 36.87 {\tiny $\pm$ 13.75} & 11.27 {\tiny $\pm$ 14.91} & 77.27 {\tiny $\pm$ 11.90} & 70.20 {\tiny $\pm$ 18.03}  \\
        \cmark & $x$ & \textbf{67.20} {\tiny $\pm$ 18.97} & \textbf{51.53} {\tiny $\pm$ 18.48} & \textbf{79.07} {\tiny $\pm$ 10.96} & \textbf{95.67} {\tiny $\pm$ 6.26 }\\
        \bottomrule
    \end{tabular}
    }
\vspace{-0.2in}
\end{table}

\subsection{Evaluation on the Offline PbRL Benchmark}
We compare LiRE with the baselines mainly on the Meta-World \texttt{medium-replay} dataset.
\cref{tab:baselines} summarizes the results of offline RL performance using ground-truth (GT) rewards and preference feedbacks respectively.
For many tasks, such as \textit{button-press-topdown} and \textit{box-close}, MR performs poorly compared to training with GT rewards, even with $1000$ preference feedbacks.
The problem of poor performance remains even if we replace the reward model with a more complex transformer architecture, PT.
PT improves performance in \textit{dial-turn} and \textit{lever-pull} tasks, but for other tasks, the performance worsens.
OPRL generally performs better than MR  due to the increased consistency of the reward models, but the performance improvement is small.
Lastly, DPPO and IPL perform better than MR on only a few tasks.
We note that existing offline PbRL methods are rarely better than MR when validated on our new dataset.

In contrast, LiRE shows a significant performance improvement over MR except for the \textit{sweep} task.
We demonstrate the importance of RLT and the linear score function by achieving high performance even when compared to SeqRank, which is also not an independent pairwise method.
In addition, policies trained with preference data outperform policies trained with GT rewards on the \textit{button-press-topdown-wall} task, suggesting that reward models trained with preference data may be more effective, as also reported in prior works \cite{christiano2017deep, kim2022preference, an2023direct}.
The results of the Meta-World \texttt{medium-expert} dataset and full learning curves are shown in the \cref{appendix:further}.

\begin{figure}
  \subfigure[MR w/ exp]{\includegraphics[width=0.44\linewidth]{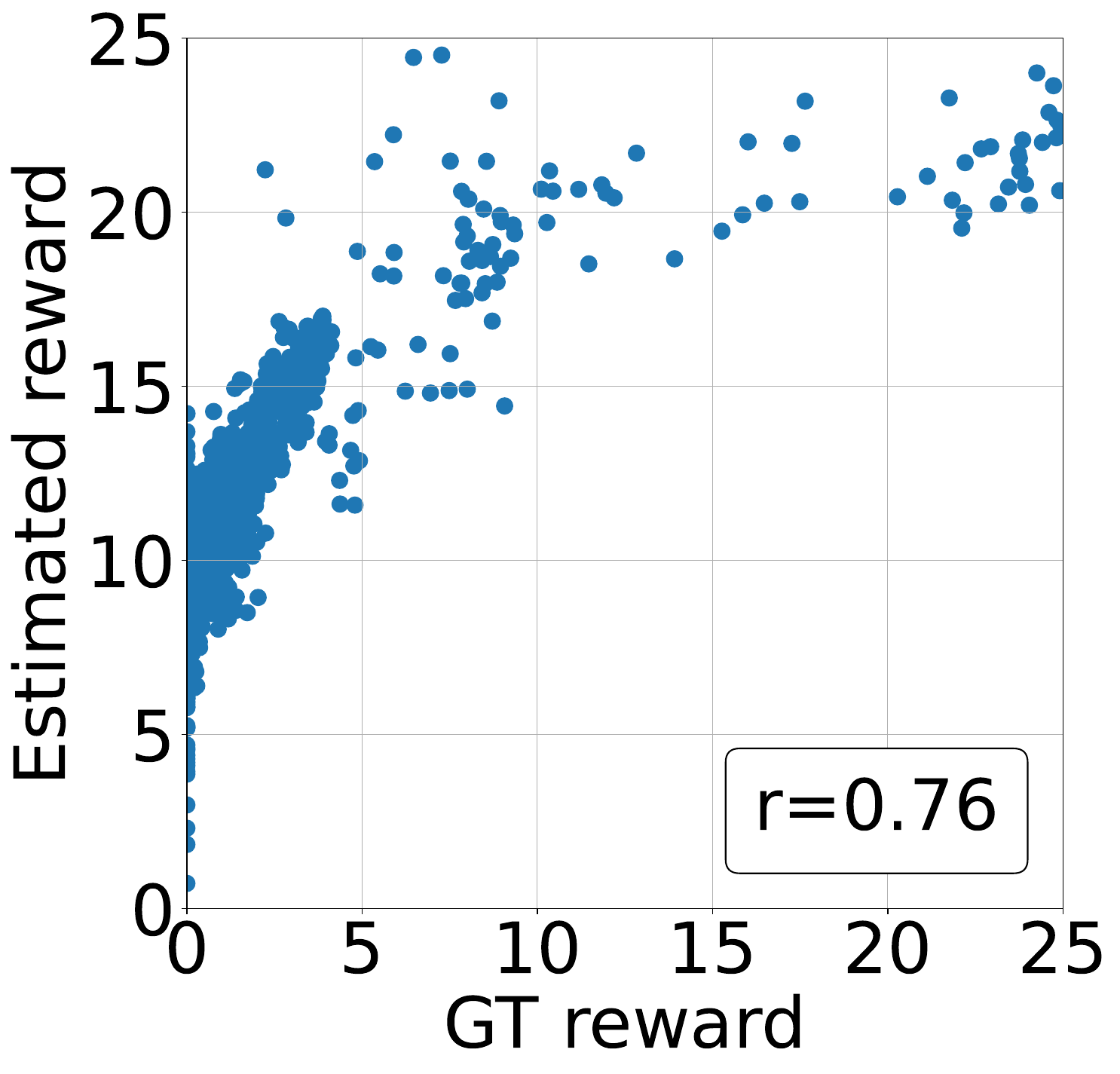}\label{fig:sub1}}
  \hfill
  \subfigure[LiRE w/ exp]{\includegraphics[width=0.44\linewidth]{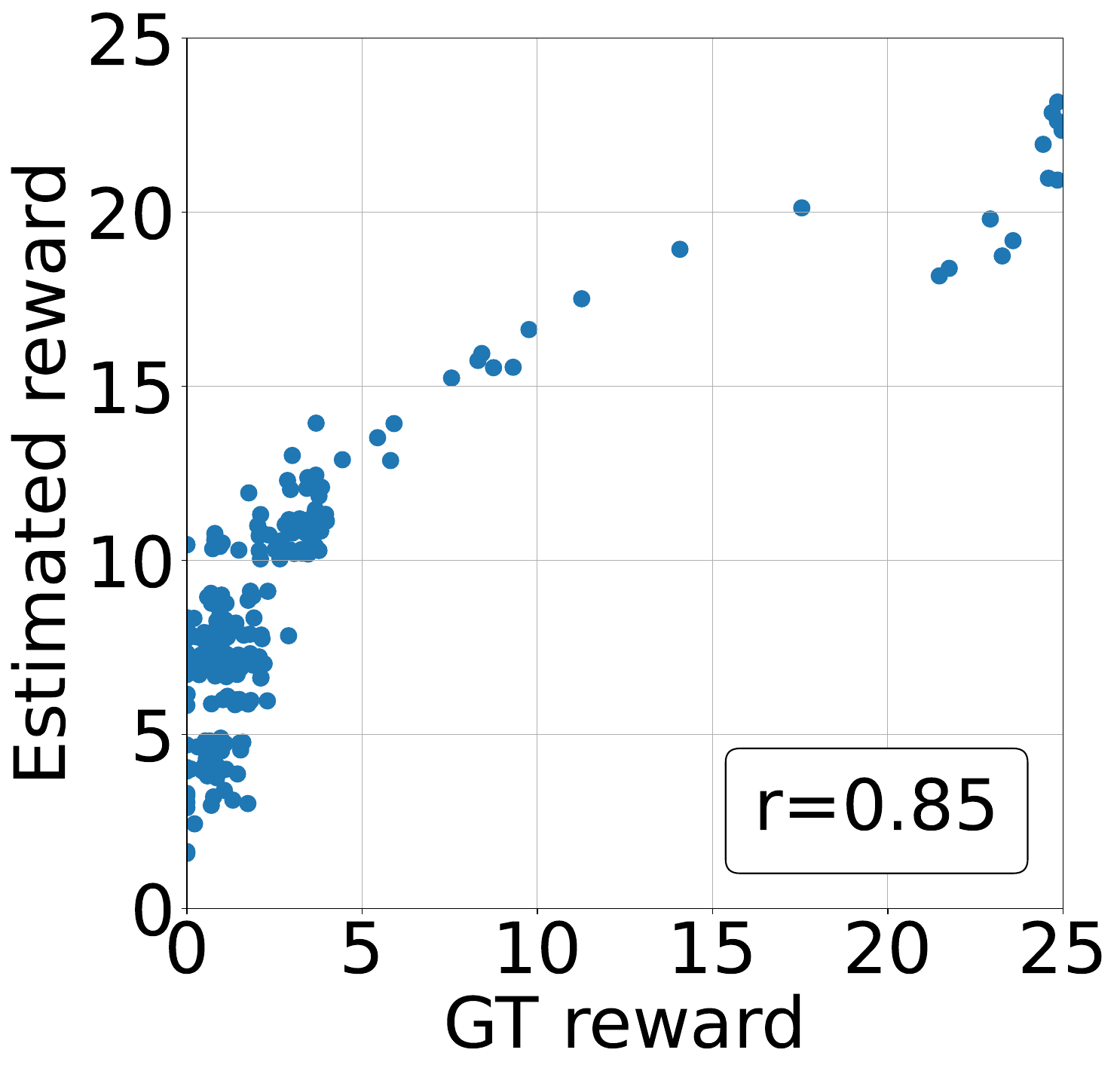}\label{fig:sub2}}
  \vspace{-0.15in}
  \vfill
  \subfigure[MR w/ linear]{\includegraphics[width=0.44\linewidth]{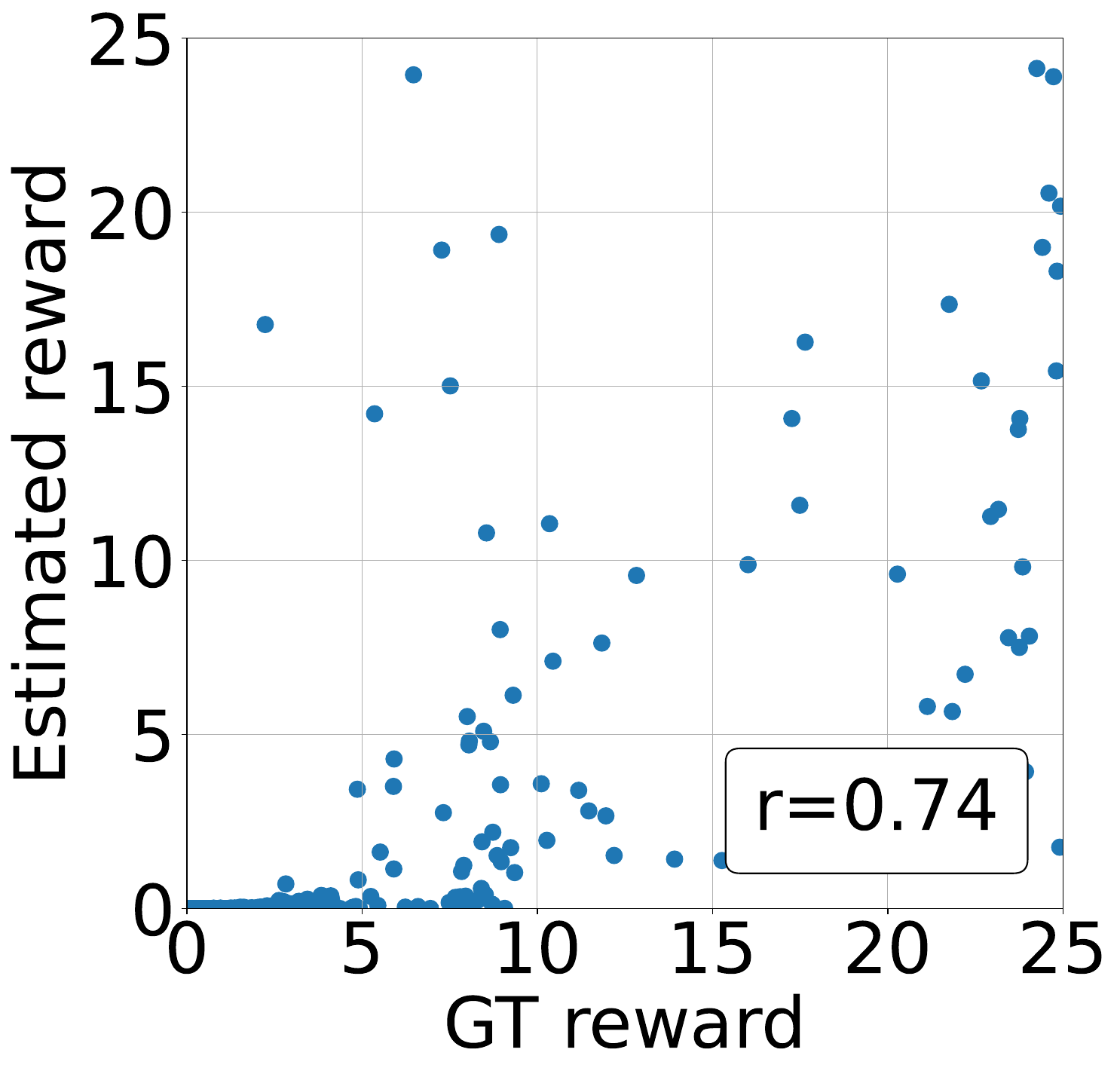}\label{fig:sub3}}
  \hfill
  \subfigure[LiRE w/ linear]{\includegraphics[width=0.44\linewidth]{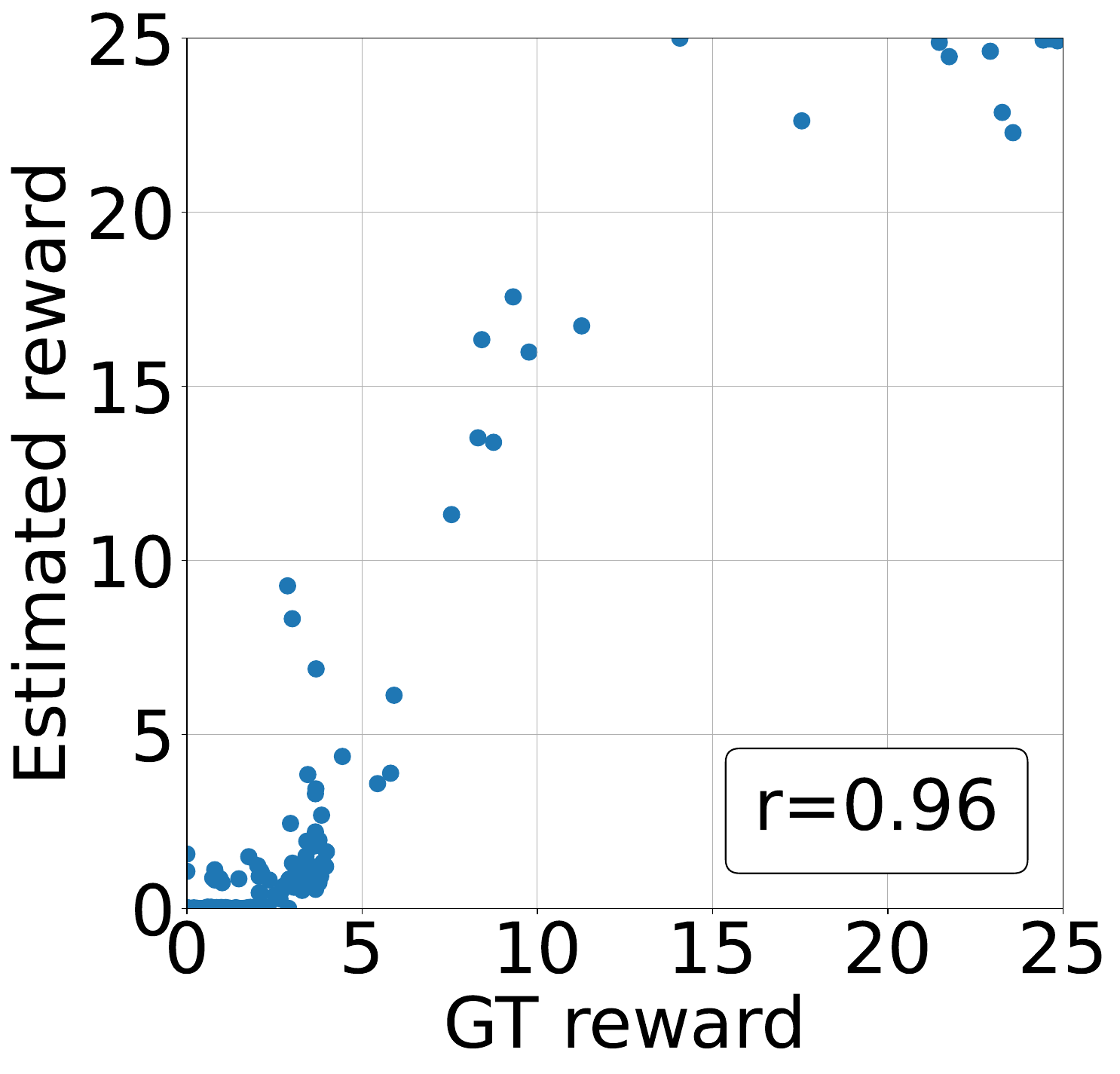}\label{fig:sub4}}
  \captionsetup{skip=0mm}
  \caption{Scatter plots of the estimated rewards for the segments used for \textit{box-close} task. The reward models are trained with MR or LiRE using the exp or linear score function. The Pearson correlation coefficient, $r$, is presented.}
  \label{fig:reward}
  \vspace{-.2in}
\end{figure}


\begin{figure*}[t]
  \centering
  \includegraphics[width=1.0\textwidth]{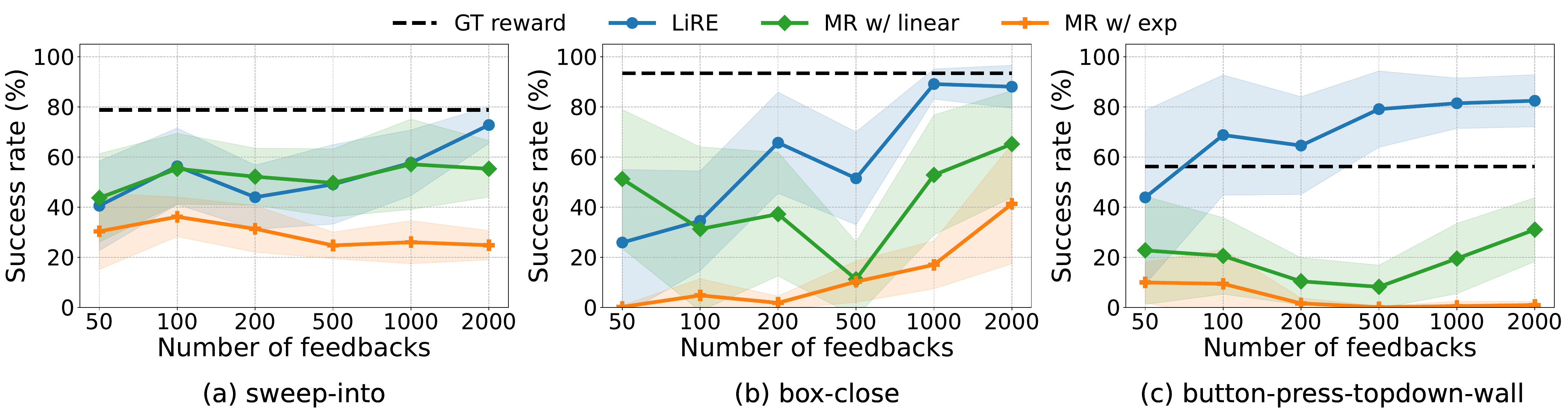}
  \captionsetup{skip=1mm}
  \caption{
Average success rates of each method while varying the number of preference feedbacks. The black dotted line represents the average success rates when trained with GT reward.
}
  \label{fig:feedback}
  \vspace{-.1in}
\end{figure*}
\begin{table*}[!ht]
	\centering
	\small
	\caption{Average success rates of LiRE when adjusting the $Q$ budget. We use a total of $500$ preference feedbacks.}
	\label{tab:q_budget}
	\begin{adjustbox}{max width=0.9\textwidth}
		\begin{tabular}{c|ccccccc|c}
			\toprule
			\multirow{2}{*}{\textbf{Dataset}} & \multirow{2}{*}{\makecell[c]{\textbf{Q=1} \\ (\textbf{MR w/ linear})}} & \multirow{2}{*}{\textbf{Q=2}} & \multirow{2}{*}{\textbf{Q=10}} & \multirow{2}{*}{\textbf{Q=20}} & \multirow{2}{*}{\textbf{Q=50}} & \multirow{2}{*}{\textbf{Q=100}} & \multirow{2}{*}{\textbf{Q=500}} & \multirow{2}{*}{\makecell[r]{\textbf{\makecell{SeqRank \\ w/ linear}}}} \\
			& & & & & & & & \\
			\midrule
			\midrule
\multirow{1}{*}{button-press-topdown} &  36.87 {\tiny$\pm$ 13.75} & 59.47 {\tiny$\pm$ 2.18} & 53.60 {\tiny$\pm$ 10.82} & 65.13 {\tiny$\pm$ 14.24} & \second{71.26} {\tiny$\pm$ 12.95} & 67.20 {\tiny$\pm$ 18.97} & \best{77.67} {\tiny$\pm$ 18.13} & \multicolumn{1}{r}{54.87 {\tiny$\pm$ 9.89}} \\
\midrule
\multirow{1}{*}{lever-pull} & 70.20 {\tiny$\pm$18.03} & 69.80 {\tiny$\pm$ 3.79} & 70.47 {\tiny$\pm$18.19} & 92.7 {\tiny$\pm$7.16} & 93.4 {\tiny$\pm$7.90} & \second{95.67} {\tiny$\pm$ 6.26} & \best{99.33} {\tiny$\pm$ 1.18} & \multicolumn{1}{r}{74.33 {\tiny$\pm$ 18.50}} \\
		\bottomrule
		\end{tabular}
	\end{adjustbox}
\vspace{-.2in}
\end{table*}

\subsection{Ablation Studies of LiRE}
\subsubsection{Factors of performance improvement}
We conduct an ablation study to verify if the performance improvement of LiRE is due to two factors: the linear score function and the RLT construction.
\cref{tab:linear} demonstrates that using the linear score function (bottom two rows) clearly brings a significant performance improvement compared to the default exponential score function (first two rows).
Additionally, the bottom two rows of \cref{tab:linear} show that constructing RLT improves performance by constructing the preference list and exploiting the second-order information.
In particular, using the linear score function with RLT has a synergistic effect, resulting in even greater improvement.

\subsubsection{Effect of RLT and score function on reward estimation}
We examine the estimated reward values of the learned reward models.
\cref{fig:reward} scatter plots the estimated rewards ($y$-axis), learned with $500$ preference feedbacks, of the segments in \textit{box-close} task against the GT rewards ($x$-axis).
Note our LiRE uses fewer segments to train the reward model, so \cref{fig:sub2} contains fewer dots than \cref{fig:sub1}.
Each segment has a length of 25 and both GT and the estimated rewards are normalized to values between $[0,25]$.

From the figure, we clearly observe that the estimated rewards in \cref{fig:sub2} are more highly correlated than those in \cref{fig:sub1}.
Namely, by constructing the RLT, LiRE exploits the second-order preference, and the high and low reward segments are more clearly distinguished by the reward estimates than vanilla MR.
Additionally, when training the reward model with the linear score function, there is a larger gap in the estimated rewards within the reward region for higher GT rewards, as shown in \cref{fig:sub4}.
We speculate that using the linear score function and RLT makes the estimated reward discern the optimal and suboptimal segments (with respect to the GT rewards) more clearly, hence, the policy learned with the estimated reward turns out to perform much better.

\subsection{Additional Analyses of LiRE}
\subsubsection{Varying the number of feedbacks}

We evaluate how the performances of the offline PbRL algorithms are affected by the number of feedbacks. Namely, we measure the average success rate of the \textit{sweep-into}, \textit{box-close}, and \textit{button-press-topdown-wall} tasks of the \texttt{medium-replay} dataset while varying the number of the preference feedbacks from $50$ to $2000$.
We note that most previous works \cite{kim2022preference, hejna2023inverse, an2023direct} using D4RL only use up to $500$ preference feedbacks.
As shown in \cref{fig:feedback}, we observe that the typical baseline, MR with exponential score function (denoted as MR w/ exp), cannot achieve a success rate higher than $50\%$ for all three tasks even with $2000$ preference feedbacks.
When we instead use the linear score function, we observe that MR w/ linear performs much better than MR w/ exp, but the success rates sometimes still remain to be low (\textit{e.g., } \textit{box-close} with 500 feedbacks and \textit{button-press-topdown-wall} for most of the times).
In contrast, it is evident that LiRE mostly surpasses the two baselines with large margins, even with fewer number of preference feedbacks. Specifically, for the \textit{button-press-topdown-wall} task, LiRE with only 100 feedbacks outperforms not only the baselines with 2000 feedbacks but also the policy learned using the GT reward. Again, we can confirm that the high feedback efficiency enabled by RLT makes LiRE very effective even with a smaller number of feedbacks.

\subsubsection{Varying Q budget}
In Section \ref{Q_budget}, we described that multiple RLTs can be constructed by putting the budget limit ($Q$) in order to increase the sample diversity. In this subsection, we show the effect of $Q$. 
\cref{tab:q_budget} shows the performance change of LiRE when varying the $Q$ budget to $1, 2, 10, 20, 50, 100,$ and $500$ while setting the total number of preference feedbacks to $500$.
Hence, for example, $Q=100$ results in five lists, and $Q=500$ results in a single list. \cref{tab:q_budget} also shows the result of SeqRank (with linear score function). From the table, we observe that since the utilization of the second-order information increases with higher values of $Q$,  the offline PbRL performance correspondingly improves, as expected. 
We also note that the performance of SeqRank is similar to that of LiRE with $Q=2$ since SeqRank creates approximately 2.3 groups in the ranked list, as detailed in \cref{tab:seq}.
This result indicates that SeqRank does not fully utilize second-order preference due to only building partially-ranked lists. A more in-depth comparison with SeqRank is given in Section \ref{sec:seqrank}.

\subsubsection{Robustness to feedback noise}
\label{exp:noise}

\begin{figure}[!t]
  \centering
  \includegraphics[width=1.0\columnwidth]{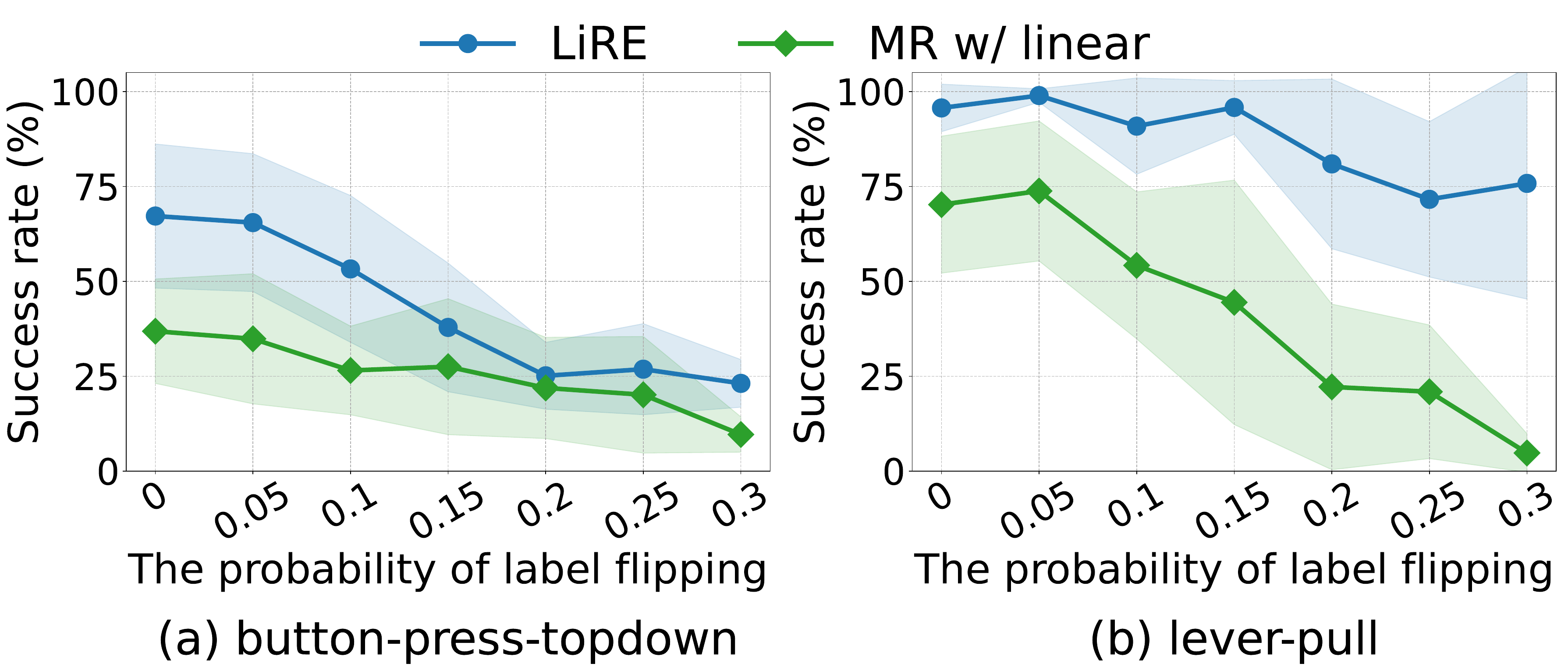}
  \caption{Robustness of LiRE w.r.t the feedback noise.}
  \label{fig:noise}
  \vspace{-.2in}
\end{figure}

If the preference feedback used in PbRL models human preference labeling, it would be reasonable to assume that the preference feedback may be noisy.
To that end, we experiment to assess the robustness of the offline PbRL performance of LiRE with respect to the preference feedback noise. 
We assume that the preference feedback can be noisy with probability $p$ (\textit{i.e., } if $l_i=0$ or $1$, the label is flipped to $1-l_i$ with probability $p$, and for tie labels, we flip to $l_i=0$ or $1$ with probability $p/2$, respectively).
We varied the noise probability $p$ from $0$ to $0.3$, and \cref{fig:noise} compares the success rates of MR w/ linear and LiRE. From the figure, we confirm that the performance of LiRE does not drop as severely as MR w/ linear when $p$ increases. 
In particular, for \textit{lever-pull} task, we observe that LiRE with feedback noise of $p=0.3$ even results in a higher success rate than MR w/ linear with no noise, highlighting the robustness of LiRE with respect to feedback noise. 


\begin{figure}[!t]
  \centering
  \includegraphics[width=1.0\columnwidth]{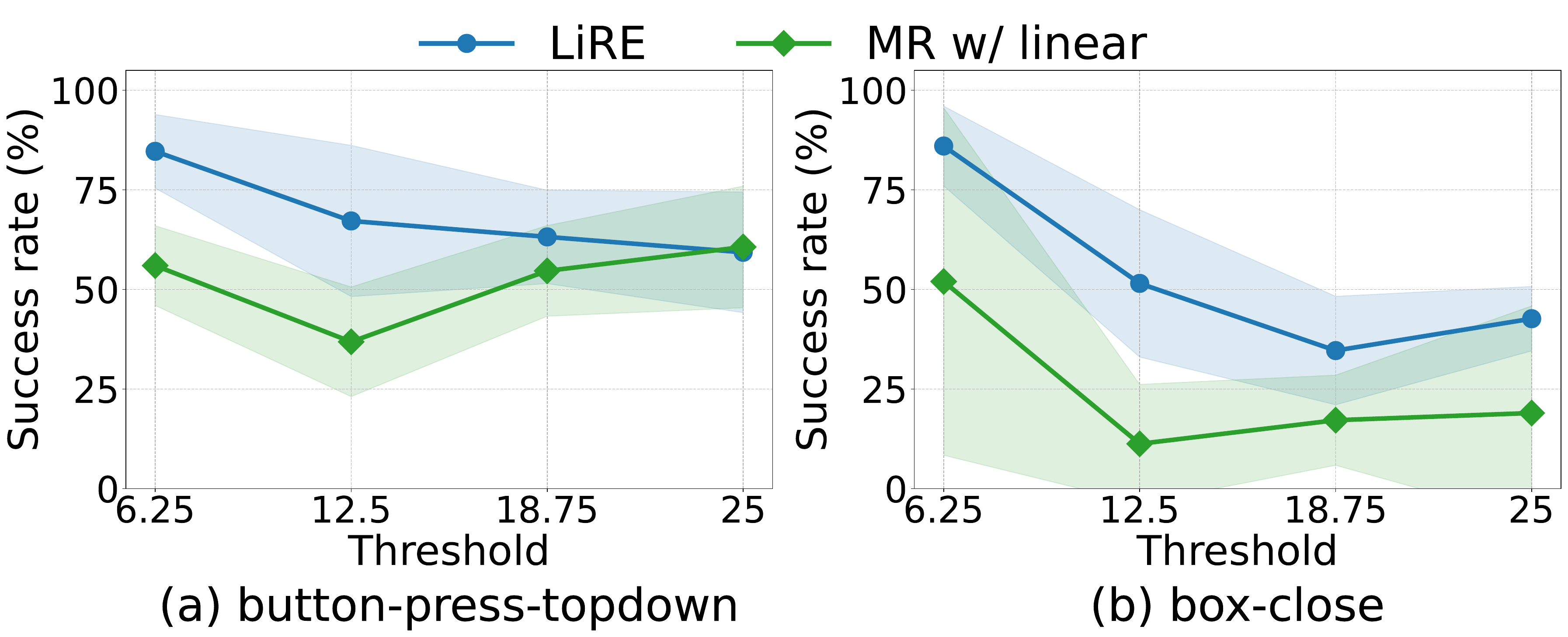}
  \caption{Effect of the granularity of preference feedback.}
  \label{fig:granular}
  \vspace{-.1in}
\end{figure}
\subsubsection{Impact of feedback granularity}

In \cref{fig:granular}, we compare the performance of LiRE based on the threshold that determines the \textit{tie} between the segments.
Specifically, we adjust the threshold value for the reward difference that indicates whether two segments are \textit{equally preferred}.
Namely, a higher threshold value means that more segment pairs are labeled as equally preferred, resulting in less granular preference feedback.
We note that the threshold value used in \cref{tab:baselines} is $12.5$ (see \cref{appendix-details} for details).
\cref{fig:granular} shows that using a smaller threshold (\textit{i.e.,} more granular feedback) improves the performance of LiRE, while the performance becomes similar to that of MR w/ linear (\textit{e.g., } \textit{button-press-topdown} task with threshold 25) when the threshold increases. Thus, we confirm that the more granular preference labels generate additional second-order preference information, which would positively affect the performance of LiRE.

\begin{table}[!t]
	\centering
	\small
	\caption{Comparison results between SeqRank and LiRE on Meta-World \texttt{medium-replay} dataset. 
 }
	\label{tab:seq}
	\resizebox{\columnwidth}{!}{
		\begin{tabular}{c|l|cccc}
			\toprule
			\multirow{2}{*}{\textbf{\makecell{\# of  \\ feedbacks}}} & \multirow{2}{*}{\textbf{Algorithm}} & \multirow{2}{*}{\textbf{\makecell[c]{Avg \\ success rate}}} & \multirow{2}{*}{\textbf{\makecell[c]{\# of  groups \\ in the list}}} & \multirow{2}{*}{\textbf{\makecell[c]{Feedback \\ efficiency}}}  & \multirow{2}{*}{\textbf{\makecell[c]{Sample \\ diversity}}}  \\
			& & \\
			\midrule
			\midrule
\multirow{2}{*}{500} & SeqRank w/ linear      & 61.84 {\tiny$\pm$ 15.80}      & 2.3 {\tiny$\pm$ 0.09}& 2.20 {\tiny$\pm$ 0.78}& \textbf{1.002} \\
& LiRE          & \best{74.83 {\tiny$\pm$ 12.23}}    & \best{9.3} {\tiny$\pm$ 1.83} & \best{11.33} {\tiny$\pm$ 3.39 }& 0.474 {\tiny$\pm$ 0.069} \\
\midrule
\multirow{2}{*}{1000} & SeqRank w/ linear   & 67.49 {\tiny$\pm$ 13.56} & 2.3 {\tiny$\pm$ 0.09}& 2.18 {\tiny$\pm$ 0.75}& \textbf{1.001} \\
& LiRE          & \best{82.92 {\tiny$\pm$ 6.48}} & \best{9.3} {\tiny$\pm$ 1.84}& \best{11.33} {\tiny$\pm$ 3.28}& 0.474 {\tiny$\pm$ 0.067} \\
		\bottomrule
		\end{tabular}
	}
\vspace{-.2in}
\end{table}

\subsubsection{Comparison with SeqRank}\label{sec:seqrank}
Here, we compare LiRE with SeqRank, which also utilizes partially-ranked lists.
We also employed the linear score function for SeqRank since it gave better results than using the exponential function and led to a fair comparison with LiRE.
We evaluate the average success rates of SeqRank and LiRE on the Meta-World \texttt{medium-replay} dataset.
The experimental results in \cref{tab:seq} show that LiRE clearly achieves higher performance than SeqRank.
We argue that SeqRank does not fully utilize the second-order information because SeqRank does not construct a fully-ranked list.
Indeed, the third column of \cref{tab:seq} shows that the number of groups in the ranked lists averages less than 3 with the SeqRank, whereas it increases to about 9 on average with LiRE.
The last two columns of \cref{tab:seq} compare feedback efficiency and sample diversity.
LiRE achieves a sample diversity of approximately 0.47 through the use of binary search, and the feedback efficiency increases significantly to 11.33 by constructing RLT.
Additionally, \cref{tab:dmc} shows the superiority of LiRE over SeqRank on the DMControl \texttt{medium-replay} dataset.
We note SeqRank also performs similarly to MR w/ linear on \textit{walker-walk} and \textit{humanoid-walk} tasks, while LiRE achieves much higher performance gains on all three tasks.
Thus, we confirm that constructing RLT and leveraging second-order preference is effective for locomotion as well as manipulation tasks.

\begin{table}[!t]
	\centering
	\small
        \captionsetup{skip=1mm}
	\caption{Average episode returns of MR, SeqRank, and LiRE on DMControl \texttt{medium-replay} dataset. We use $500$ preference feedbacks.}
	\label{tab:dmc}
	\resizebox{\columnwidth}{!}{
		\begin{tabular}{l|rrr}
			\toprule
			{\textbf{Algorithm}} & {\textbf{\makecell{hopper-hop}}} & {\textbf{\makecell{walker-walk}}} & {\textbf{\makecell{humanoid-walk}}}  \\
			\midrule
			\midrule
IQL with GT rewards      &   157.95 {\tiny$\pm$ 9.64}     &  839.6  {\tiny$\pm$ 36.57}    &   250.9 {\tiny$\pm$ 11.62} \\
			\midrule
MR w/ linear     & 53.96 {\tiny$\pm$ 24.42}      & 677.38 {\tiny$\pm$ 88.14} & 84.35 {\tiny$\pm$ 23.23}  \\
SeqRank w/ linear          &  80.84 {\tiny$\pm$ 27.67}   & 698.81 {\tiny$\pm$ 91.71}  & 80.68 {\tiny$\pm$ 14.67}\\
LiRE          & \best{99.14} {\tiny$\pm$ 12.28}    & \best{822.27} {\tiny$\pm$ 50.83}  & \best{104.08} {\tiny$\pm$ 17.45}  \\
		\bottomrule
		\end{tabular}
	}
\vspace{-.2in}
\end{table}

\subsubsection{Compatibility with other methods }
\begin{figure}[!t]
  \subfigure[LiRE with OPRL]{\includegraphics[width=0.48\linewidth]{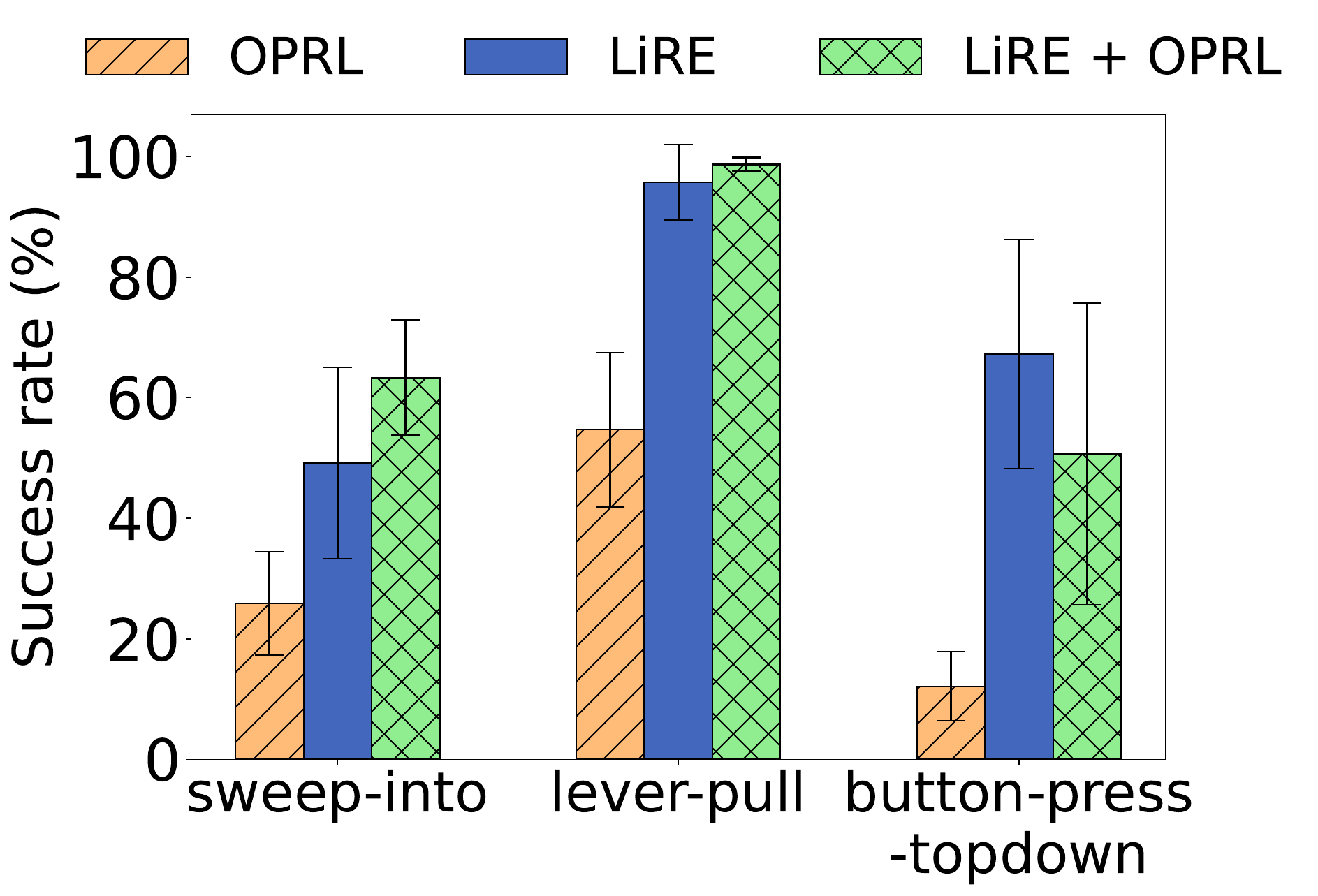}\label{fig:OPRL_sub1}}
  \subfigure[LiRE with PT]{\includegraphics[width=0.48\linewidth]{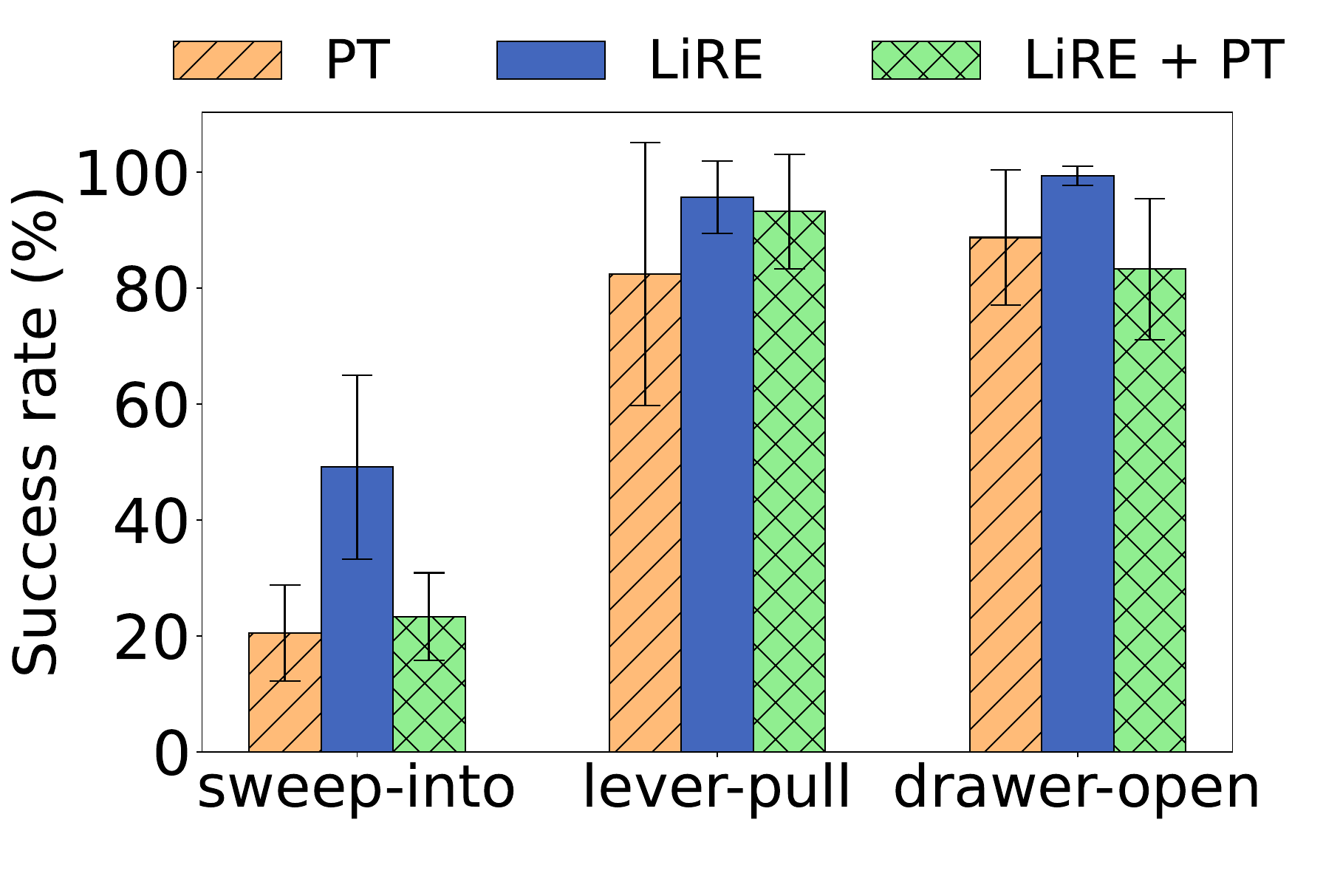}\label{fig:PT_sub2}}
  \vspace{-0.15in}
  \vfill
  \caption{Combining LiRE with other baselines.}
  \label{fig:others_plus_LiRE}
  \vspace{-.2in}
\end{figure}

To check the compatibility of LiRE with other methods, we tested the performance of LiRE when combined with OPRL and PT, respectively. 
First, to apply OPRL and LiRE simultaneously, we trained a reward model each time an RLT was newly constructed, and then actively sampled based on the disagreement of the reward models (following the method of OPRL) when constructing the next RLT.
In \cref{fig:OPRL_sub1}, we observe that LiRE+OPRL outperforms LiRE in \textit{sweep-into} and \textit{lever-pull} tasks but performs worse in \textit{button-press-topdown} task.
This discrepancy suggests that while the OPRL method enhances the consistency of the reward model, it may lead to oversampling similar segments that are challenging to distinguish depending on the task.
Second, as shown in \cref{fig:PT_sub2}, LiRE does not necessarily gain improvements when combined with PT. That is, since PT was originally designed to capture temporal dependencies of segments in reward modeling, it seems to struggle in accurately capturing the second-order preference information from RLT possibly due to overfitting to the sequence of past segments.

\subsection{Human Experiments}

\label{exp:human}

\cref{tab:human} presents the results with \textit{real} human preference feedback on the new \textit{button-press-topdown} offline RL dataset, which is distinct from the dataset used in \cref{tab:baselines}.
Namely, we collected $200$ preference feedbacks from one of the authors for each of the three feedback collection methods: MR, SeqRank, and LiRE.
For LiRE, we used the feedback budget of $Q=100$, resulting in two RLTs.
The results again indicate that LiRE dominates other baselines and gets stronger when the linear score function is used. We believe this result shows the potential of LiRE that it can be very effective in practical scenarios with real human preference feedback, as in LLM alignment.

\begin{table}[!t]
	\centering
	\small
        \captionsetup{skip=1mm}
	\caption{Average success rates of a \textit{button-press-topdown} task with $200$ of real human feedbacks. }
	\label{tab:human}
	\resizebox{0.7\columnwidth}{!}{
		\begin{tabular}{c|ccc}
			\toprule
			{\textbf{$\phi(x)$}} & \textbf{\makecell{MR}} & \textbf{\makecell{SeqRank}} & \textbf{\makecell{LiRE}}  \\
			\midrule
			\midrule
$\text{exp}(x)$      & 38.00 {\tiny$\pm$ 8.85}     & 40.93 {\tiny$\pm$ 10.72} & 	\second{78.27} {\tiny$\pm$ 6.59} \\
			\midrule
$x$         &  43.33 {\tiny$\pm$ 25.72}  & 74.13 {\tiny$\pm$ 9.96} & \best{90.67} {\tiny$\pm$ 7.57}\\
		\bottomrule
		\end{tabular}
	}
\vspace{-.2in}
\end{table}

\section{Limitation}
We believe there are two limitations of LiRE.
First, LiRE lacks the ability to parallelize the construction of RLT since there are dependencies between the order in which feedbacks are obtained to construct a fully-ranked list.
Therefore, in scenarios where parallel feedback collection is feasible, constructing an RLT could be more time-consuming compared to collecting preference feedbacks independently in pairs.
Nevertheless, the results presented in \cref{appendix:200feedback} show that LiRE with only $200$ feedbacks outperforms the independent pairwise sampling method using $1000$ feedbacks, suggesting the importance of constructing RLT. Second, LiRE relies on the transitivity assumption outlined in Assumption \ref{assumption}.
Although our experiments with feedback noise indicate LiRE's robustness to noise that violates this assumption, transitivity violations can occur even with noiseless labels in real-world applications.
This issue is not unique to LiRE but affects other preference-based RL methods as well.
Addressing transitivity violation remains a challenge for scalar reward models, so future research could explore solutions by using multi-dimensional preference feedback to construct RLTs for each dimension.

\section{Concluding Remarks}
In this paper, we propose a novel Listwise Reward Estimation (LiRE) method for offline preference-based RL. 
While obtaining second-order preference from a traditional framework is challenging, we demonstrate that LiRE efficiently exploits second-order preference by constructing an RLT using ordinary, simple ternary feedback.
Our experiments demonstrate the significant performance gains achieved by LiRE on our new offline PbRL dataset, specifically designed to objectively reflect the effect of estimated rewards.
Notably, the reward model trained with LiRE outperforms traditional pairwise feedback methods, even with fewer preference feedbacks, highlighting the importance of second-order preference information.
Moreover, our findings suggest that constructing ranked lists can be straightforward without complex second-order preference feedback, indicating the broad applicability of LiRE to more challenging tasks and real-world applications.
\section*{Impact Statement}
We believe our LiRE can be potentially applied to aligning the RL agent with more fine-grained human intent and preference. Such applications can bring significant societal consequences by enhancing the precision and effectiveness of AI systems in various fields such as health care and education. By ensuring that AI systems more closely reflect and respond to the detailed intentions of their users, LiRE has the potential to foster trust and acceptance of AI technologies, ultimately contributing to their more widespread and ethical adoption.

\section*{Acknowledgments}

This work was supported in part by the National Research Foundation of Korea (NRF) grant [No.2021R1A2C2007884] and by Institute of Information \& communications Technology Planning \& Evaluation (IITP) grants
[RS-2021-II211343, RS-2021-II212068, RS-2022-II220113,
RS-2022-II220959] funded by the Korean government (MSIT). It was also supported by AOARD Grant No. FA2386-23-1-4079. 

\bibliography{main_reference}
\bibliographystyle{icml2024}

\newpage
\appendix
\onecolumn
\section{Additional Experimental Analysis of LiRE}
\label{appendix:further}

\subsection{Experimental Results on \texttt{medium-expert} Dataset}

We summarize the experimental results on the Meta-World \texttt{medium-expert} dataset in \cref{tab:medium-expert results}.
LiRE outperforms significantly baselines for \textit{sweep} and \textit{hammer} tasks.
While DPPO performs better than LiRE in the case of \textit{box-close} task, DPPO performs poorly compared to basic MR in other tasks.

\begin{table*}[ht]
	\centering
	\small
	\caption{Average success rate of the algorithms on Meta-World \texttt{medium-expert} dataset over six random seeds. We use $500$ and $1000$ preference feedbacks and LiRE significantly outperforms the existing baselines.}
	\label{tab:medium-expert results}
	\begin{adjustbox}{max width=0.5\textwidth}
		\begin{tabular}{c|l|rrr}
			\toprule
			\multirow{2}{*}{\textbf{\# of feedback}} & \multirow{2}{*}{\textbf{Algorithm}} & \multirow{2}{*}{\textbf{box-close}} & \multirow{2}{*}{\textbf{sweep}} & \multirow{2}{*}{\textbf{hammer}}  \\
			& & & \\
			\midrule
			\midrule
-  & IQL with GT rewards & 65.00 \tiny{$\pm$} 9.98 & 85.33 \tiny{$\pm$} 5.96 &  65.00 \tiny{$\pm$} 11.36 \\ \cmidrule{1-5}
\multirow{7}{*}{500} & MR          & 15.33 \tiny{$\pm$} 6.99 & 42.67 \tiny{$\pm$} 9.21 & 5.67 \tiny{$\pm$} 6.97   \\
& PT \cite{kim2022preference}         & 2.33 \tiny{$\pm$} 3.54 & 57.33 \tiny{$\pm$} 8.92 & 1.67 \tiny{$\pm$} 2.92  \\
& OPRL  \cite{shin2022benchmarks}      & 10.00 \tiny{$\pm$} 9.87 & 30.00 \tiny{$\pm$} 8.87 & 6.00 \tiny{$\pm$} 4.62   \\
& DPPO \cite{an2023direct}       & 41.00 \tiny{$\pm$} 9.50 & 12.33 \tiny{$\pm$} 8.44 & 8.00 \tiny{$\pm$} 10.07  \\
& IPL \cite{hejna2023inverse}        & 7.00 \tiny{$\pm$} 9.50 & 15.33 \tiny{$\pm$} 10.37 &  4.33 \tiny{$\pm$} 4.23 \\
& SeqRank \cite{hwang2023sequential}        & 10.73 \tiny{$\pm$} 7.58 & 53.8 \tiny{$\pm$} 18.21 &  4.6 \tiny{$\pm$} 5.27 \\
\cmidrule{2-5}
& LiRE (ours) & 25.26 \tiny{$\pm$} 11.70 & 59.53 \tiny{$\pm$} 26.92 & 50.20 \tiny{$\pm$} 16.98  \\
\midrule
\multirow{7}{*}{1000} & MR          & 13.67 \tiny{$\pm$} 11.57 & 50.00 \tiny{$\pm$} 8.64 & 8.00 \tiny{$\pm$} 8.25   \\
& PT  \cite{kim2022preference}        & 13.00 \tiny{$\pm$} 12.26 & 21.00 \tiny{$\pm$} 15.44 & 1.33 \tiny{$\pm$} 2.98  \\
& OPRL \cite{shin2022benchmarks}       & 11.33 \tiny{$\pm$} 6.80 & 44.33 \tiny{$\pm$} 6.67 & 6.33 \tiny{$\pm$} 5.47  \\
& DPPO  \cite{an2023direct}      & 42.67 \tiny{$\pm$} 15.52 & 14.33 \tiny{$\pm$} 13.19 &  5.33 \tiny{$\pm$} 4.85 \\
& IPL \cite{hejna2023inverse}        & 10.67 \tiny{$\pm$} 6.90 & 16.67 \tiny{$\pm$} 11.64 & 9.33 \tiny{$\pm$} 9.14  \\
& SeqRank \cite{hwang2023sequential}        & 13.4 \tiny{$\pm$} 8.89 & 68.45 \tiny{$\pm$} 14.26 &  21.27 \tiny{$\pm$} 17.86 \\
\cmidrule{2-5}
& LiRE (ours) & 27.53 \tiny{$\pm$} 20.45 & 73.00 \tiny{$\pm$} 25.68  & 41.66 \tiny{$\pm$} 32.64 \\
		\bottomrule
		\end{tabular}
	\end{adjustbox}
\end{table*}

\subsection{LiRE with Fewer Preference Feedbacks}

If we have pre-collected independent pairwise preference data, MR can use the entire preference data.
However, LiRE has the disadvantage of requiring additional feedbacks between segments for constructing RLT.
Nevertheless, \cref{tab:200_feedback} shows the importance of RLT to obtain second-order preference.
The performance of LiRE with $200$ feedbacks is better than the performance using $1000$ independent pairwise feedbacks.

\label{appendix:200feedback}
\begin{table*}[h]
	\centering
	\small
	\caption{Average success rates on Meta-World \texttt{medium-replay} dataset. We use $1000$ preference feedbacks for MR and $200$ preference feedbacks for LiRE.}
	\label{tab:200_feedback}
	\begin{adjustbox}{max width=\textwidth}
		\begin{tabular}{c|l|rrrrrrrr}
			\toprule
			\multirow{2}{*}{\textbf{\makecell[c]{\# of\\ feedbacks} }} & \multirow{2}{*}{\textbf{Algorithm}} & \multirow{2}{*}{\textbf{\makecell[r]{button-press \\ -topdown}}} & \multirow{2}{*}{\textbf{box-close}} & \multirow{2}{*}{\textbf{dial-turn}} & \multirow{2}{*}{\textbf{sweep}} & \multirow{2}{*}{\textbf{\makecell[r]{button-press \\ -topdown-wall}}} & \multirow{2}{*}{\textbf{sweep-into}} & \multirow{2}{*}{\textbf{drawer-open}} & \multirow{2}{*}{\textbf{lever-pull}}  \\
			& & & & & & & \\
			\midrule
			\midrule
-  & IQL with GT rewards & 88.33 \tiny{$\pm$ 4.76 }    & 93.40 \tiny{$\pm$ 3.10 } & 75.40 \tiny{$\pm$ 5.47 } & 98.33 \tiny{$\pm$ 1.87 } & 56.27 \tiny{$\pm$ 6.32 }         & 78.80 \tiny{$\pm$ 7.96 } & 100.00 \tiny{$\pm$ 0.00 }& 98.47 \tiny{$\pm$ 1.77 } \\ \cmidrule{1-10}
1000 & MR          & 9.27 \tiny{$\pm$ 5.30 }& 17.07 \tiny{$\pm$ 9.56 }& 59.07 \tiny{$\pm$ 7.57 }& 90.80 \tiny{$\pm$ 9.74 }& 0.60 \tiny{$\pm$ 1.87 }& 26.07 \tiny{$\pm$ 8.57 }& 96.47 \tiny{$\pm$ 4.02 }& 50.87 \tiny{$\pm$ 10.89}
  \\
\midrule
200 & LiRE & 36.60 \tiny{$\pm$ 16.30}	& 60.33 \tiny{$\pm$ 23.96}	& 80.73 \tiny{$\pm$ 10.53}	& 76.87 \tiny{$\pm$ 13.86}	& 53.27 \tiny{$\pm$ 24.12}	& 39.33 \tiny{$\pm$ 10.62}	 & 99.93 \tiny{$\pm$ 0.36 }& 	81.47 \tiny{$\pm$ 13.53}  \\
		\bottomrule
		\end{tabular}
	\end{adjustbox}
\vspace{-.2in}
\end{table*}

\subsection{Training LiRE with Listwise Loss}
\label{appendix:listwise}

\cref{method:LiRE} describes how to train the reward model with pairwise loss from constructed RLT.
However, we can apply listwise loss in addition to pairwise loss since a ranked list is constructed.
In this section, we introduce how to train the reward model with listwise loss.
Suppose that we have $n$ segments, $(\sigma_1, \sigma_2, \cdots, \sigma_n)$ and denote the rewards of the segments, $\bigl(r(\sigma_1), r(\sigma_2), \cdots, r(\sigma_n)\bigr)$.
We assume the probability of permutation of $n$ segments follows a Plackett-Luce (PL) model
\cite{plackett1975analysis}:
\begin{equation}
    P(\bm{\pi}) = \prod_{i=1}^{n}\frac{\phi\bigl(r(\sigma_{\pi_i})\bigr)}{\sum_{j=i}^{n}\phi\bigl(r(\sigma_{\pi_j})\bigr)}
    \label{eq:list eq}
\end{equation}
where $\phi$ is an increasing and strictly positive function and $\bm{\pi}=(\pi_1, \pi_2, \cdots, \pi_n)$ is a permutation of $(1,2,\cdots,n)$.
Here, $P(\bm{\pi})$ is the probability distribution in which $n$ segments are ranked in order of permutation $\bm{\pi}$, indicating the likelihood of segment $\sigma_i$ being ranked $\pi_i$-th.

Since we do not know the true probability of permutation, we set the score of the segment based on the ranks.
Specifically, given $k$ ranks in the list, let $s(\sigma)=(k+1-m)R/k$ be the score of the segment $\sigma$ that belongs to the $m$-th preferred rank (\textit{i.e.,} $\sigma \in g_{k+1-m}$) where $m \in \{1, \cdots, k\}$ and $R$ is constant.
In our implementation, the constant $R$ is set to the maximum boundary of the output of the reward model, which is bounded by $[0, R]$ by the Tanh function.

Our goal is to minimize the following objective:
\begin{equation}
    D_{KL} \left( P_s(\bm{\pi}) \middle\| P_{\theta}(\bm{\pi}) \right) = D_{KL} \left( 
    \prod_{i=1}^{n}\frac{\phi(s(\sigma_{\pi_i}))}{\sum_{j=i}^{n}\phi(s(\sigma_{\pi_j}))}
    \middle\| 
    \prod_{i=1}^{n}\frac{\phi\bigl(r_{\theta}(\sigma_{\pi_i})\bigr)}{\sum_{j=i}^{n}\phi\bigl(r_{\theta}(\sigma_{\pi_j})\bigr)}
    \right).
\end{equation}

Since the number of permutations grows by $n!$, computing the permutation probability demands a high computational cost.
Thus, we minimize the following objective using the top one probability proposed by ListNet \cite{xia2008listwise}:
\begin{equation}
    \sum_{i=1}^{n} D_{KL} \left( P_s(i) \middle\| P_{\theta}(i) \right)
\end{equation}
where 
\begin{equation}
    P_s(i) =  P_s(\pi_1 = i) = \frac{\phi(s(\sigma_i))}{\sum_{j=1}^{n}\phi(s(\sigma_j))}
\end{equation}
and
\begin{equation}
    P_{\theta}(i) = P_{\theta}(\pi_1 = i) = \frac{\phi(r_{\theta}(\sigma_i))}{\sum_{j=1}^{n}\phi(r_{\theta}(\sigma_j))}.
\end{equation}

We train the reward model by sampling $n=10$ segments from the RLT at each gradient descent.
\cref{tab:listwise} compares the performance of LiRE trained with listwise and pairwise losses.
As shown in \cref{tab:listwise}, training the reward model with pairwise loss is more stable and performs better in most cases, except for \textit{sweep} and \textit{sweep-into} tasks.

\begin{table*}[h]
	\centering
	\small
	\caption{Average success rates on \texttt{medium-replay} dataset when using the listwise loss for training the reward model.}
	\label{tab:listwise}
	\begin{adjustbox}{max width=\textwidth}
		\begin{tabular}{c|l|rrrrrrrr}
			\toprule
			\multirow{2}{*}{\textbf{\makecell[c]{\# of\\ feedbacks}}} & \multirow{2}{*}{\textbf{Algorithm}} & \multirow{2}{*}{\textbf{\makecell[r]{button-press \\ -topdown}}} & \multirow{2}{*}{\textbf{box-close}} & \multirow{2}{*}{\textbf{dial-turn}} & \multirow{2}{*}{\textbf{sweep}} & \multirow{2}{*}{\textbf{\makecell[r]{button-press \\ -topdown-wall}}} & \multirow{2}{*}{\textbf{sweep-into}} & \multirow{2}{*}{\textbf{drawer-open}} & \multirow{2}{*}{\textbf{lever-pull}}  \\
			& & & & & & & \\
			\midrule
			\midrule
-  & IQL with GT rewards & 88.33 \tiny{$\pm$ 4.76}    & 93.40 \tiny{$\pm$ 3.10} & 75.40 \tiny{$\pm$ 5.47} & 98.33 \tiny{$\pm$ 1.87} & 56.27 \tiny{$\pm$ 6.32}         & 78.80 \tiny{$\pm$ 7.96} & 100.00 \tiny{$\pm$ 0.00}& 98.47 \tiny{$\pm$ 1.77} \\ \cmidrule{1-10}
\multirow{2}{*}{500} & LiRE w/ listwise & 53.13 \tiny{$\pm$ 10.63}    & 55.07 \tiny{$\pm$ 16.11} & 63.87 \tiny{$\pm$ 8.39} & 99.53 \tiny{$\pm$ 1.12} & 17.73 \tiny{$\pm$ 11.51}         & 63.47 \tiny{$\pm$ 11.47} & 98.60 \tiny{$\pm$ 3.27} & 84.53 \tiny{$\pm$ 10.33} \\ 
& LiRE w/ pairwise & 67.20 \tiny{$\pm$ 18.97}    & {51.53} \tiny{$\pm$ 18.48} & {79.07} \tiny{$\pm$ 10.96} & 77.53 \tiny{$\pm$ 10.50} & {79.13} \tiny{$\pm$ 15.19}         & {49.13} \tiny{$\pm$ 15.85} & {99.40} \tiny{$\pm$ 1.65} & {95.67} \tiny{$\pm$ 6.26} \\
\midrule
\multirow{2}{*}{1000} & LiRE w/ listwise & 55.73 \tiny{$\pm$ 8.57}& 68.07 \tiny{$\pm$ 9.06}& 68.20 \tiny{$\pm$ 9.37}& 99.07 \tiny{$\pm$ 1.44}& 23.93 \tiny{$\pm$ 7.31}& 62.60 \tiny{$\pm$ 12.21} & 99.40 \tiny{$\pm$ 2.08}& 83.80 \tiny{$\pm$ 7.97}\\
& LiRE w/ pairwise & {83.07} \tiny{$\pm$ 6.38}& {89.13} \tiny{$\pm$ 6.02}& {76.93} \tiny{$\pm$ 7.55}& 75.87 \tiny{$\pm$ 6.81}& {81.47} \tiny{$\pm$ 10.04} & {57.73} \tiny{$\pm$ 13.11} & {99.73} \tiny{$\pm$ 0.85}& {99.47} \tiny{$\pm$ 1.15} \\
		\bottomrule
		\end{tabular}
	\end{adjustbox}
\vspace{-.2in}
\end{table*}

\subsection{Increasing Epochs of Reward Model Training}
As described in \cref{appendix:hyperparameters}, the epochs experimented with in \cref{tab:baselines} is $300$.
\cref{tab:long_epochs} shows the performance when we increase the epochs to $5000$.
Both MR and LiRE tend to perform better with more epochs, but LiRE still performs better than MR.
The performance gap between using the exponential score function and the linear score function for LiRE is smaller at $5000$ epochs than at $300$ epochs.
However, when the epoch is $5000$, the linear score function has a significant performance improvement on the \textit{dial-turn} and \textit{button-press-topdown-wall} tasks and performs better or similar to the exponential score function on other tasks.

\begin{table*}[ht]
	\centering
	\small
	\caption{Average success rates on Meta-World \texttt{medium-replay} dataset with increased epochs. There is a performance improvement when training with longer epochs.}
	\label{tab:long_epochs}
	\begin{adjustbox}{max width=\textwidth}
		\begin{tabular}{c|l|rrrrrrrr}
			\toprule
			\multirow{2}{*}{\textbf{\makecell[c]{Epochs}}} & \multirow{2}{*}{\textbf{Algorithm}} & \multirow{2}{*}{\textbf{\makecell[r]{button-press \\ -topdown}}} & \multirow{2}{*}{\textbf{box-close}} & \multirow{2}{*}{\textbf{dial-turn}} & \multirow{2}{*}{\textbf{sweep}} & \multirow{2}{*}{\textbf{\makecell[r]{button-press \\ -topdown-wall}}} & \multirow{2}{*}{\textbf{sweep-into}} & \multirow{2}{*}{\textbf{drawer-open}} & \multirow{2}{*}{\textbf{lever-pull}}  \\
			& & & & & & & \\
			\midrule
			\midrule
-  & IQL with GT rewards & 88.33 \tiny{$\pm$ 4.76}    & 93.40 \tiny{$\pm$ 3.10} & 75.40 \tiny{$\pm$ 5.47} & 98.33 \tiny{$\pm$ 1.87} & 56.27 \tiny{$\pm$ 6.32}         & 78.80 \tiny{$\pm$ 7.96} & 100.00 \tiny{$\pm$ 0.00}& 98.47 \tiny{$\pm$ 1.77} \\ \cmidrule{1-10}
\multirow{3}{*}{300}  & MR &  9.60 \tiny{$\pm$ 5.74} & 10.33 \tiny{$\pm$ 8.23} &  50.20 \tiny{$\pm$ 8.51} & 79.80 \tiny{$\pm$ 13.36}  &  0.13 \tiny{$\pm$ 0.50}  &  24.80 \tiny{$\pm$ 5.28}  &  98.07 \tiny{$\pm$ 3.20} &  50.53 \tiny{$\pm$ 8.55}\\
& LiRE w/ exp &  12.87 \tiny{$\pm$ 7.86} & 22.73 \tiny{$\pm$ 10.40} & 65.87 \tiny{$\pm$ 9.46} & 82.67 \tiny{$\pm$ 19.86}  &  1.33 \tiny{$\pm$ 2.15} & 24.87 \tiny{$\pm$ 8.39}&   98.67 \tiny{$\pm$ 1.89}&  57.87 \tiny{$\pm$ 11.28}\\ 
& LiRE w/ linear & 67.20 \tiny{$\pm$ 18.97}    & {51.53} \tiny{$\pm$ 18.48} & {79.07} \tiny{$\pm$ 10.96} & 77.53 \tiny{$\pm$ 10.50} & {79.13} \tiny{$\pm$ 15.19}         & {49.13} \tiny{$\pm$ 15.85} & {99.40} \tiny{$\pm$ 1.65} & {95.67} \tiny{$\pm$ 6.26} \\
\midrule
\multirow{3}{*}{5000} & MR & 32.87 \tiny{$\pm$ 9.94}  & 31.80 \tiny{$\pm$ 12.65} & 60.33 \tiny{$\pm$ 8.34} & 93.5 \tiny{$\pm$ 6.61} &  25.40 \tiny{$\pm$ 10.25}  &  36.00 \tiny{$\pm$ 9.58} & 98.00 \tiny{$\pm$ 4.00}& 75.93 \tiny{$\pm$ 6.46}\\
& LiRE w/ exp & 68.33  \tiny{$\pm$ 16.33}  & 83.13  \tiny{$\pm$ 10.36} & 77.53  \tiny{$\pm$ 6.25} & 91.87 \tiny{$\pm$  7.02}  &  36.80  \tiny{$\pm$ 14.17}  & 59.53  \tiny{$\pm$ 14.99}   &  99.93  \tiny{$\pm$ 0.36}& 79.47 \tiny{$\pm$  8.61} \\
& LiRE w/ linear  &  77.27 \tiny{$\pm$  13.52} &  76.6  \tiny{$\pm$ 17.16} &  88.33  \tiny{$\pm$ 5.49} & 87.6 \tiny{$\pm$  15.45}  &  77.27  \tiny{$\pm$ 13.52}  &  60.67  \tiny{$\pm$ 11.96}  & 97.07  \tiny{$\pm$ 5.63} & 83.4  \tiny{$\pm$ 6.43}\\
		\bottomrule
		\end{tabular}
	\end{adjustbox}
\vspace{-.1in}
\end{table*}

\subsection{Applying a Linear Score Function to Other Baselines}
Many existing studies utilize the exponential score function for PbRL using human feedback \cite{christiano2017deep, lee2021pebble}.
Nevertheless, numerous other score functions are also prevalent in the PbRL literature such as Table 1 in the survey paper of PbRL \cite{wirth2017survey}.
Additionally, \cite{song2023preference} have demonstrated that alternative score functions are effective in RLHF.
We also present the performance of baselines using the linear score function instead of the exponential function across four Meta-World \texttt{medium-replay} tasks in \cref{tab:linear_to_baselines}.
\cref{tab:linear_to_baselines} reveals that LiRE, when utilizing the linear score function, surpasses all other baselines, even when these baselines also use the linear score function.
For PT or DPPO, there is no performance improvement when using a linear score function.
We leave it as future work to analyze which score functions are effective depending on the model structure or training method.

\begin{table*}[ht]
	\centering
	\small
	\caption{Average success rates when applying the linear and exponential score functions to the baselines.}
	\label{tab:linear_to_baselines}
	\begin{adjustbox}{max width=0.8\textwidth}
		\begin{tabular}{c|c|rrrrrr}
			\toprule
			\multirow{2}{*}{\textbf{\makecell[c]{Task} }} & \multirow{2}{*}{\textbf{\makecell[c]{$\phi(x)$}}} & \multirow{2}{*}{\textbf{MR}} & \multirow{2}{*}{\textbf{\makecell[r]{PT}}} & \multirow{2}{*}{\textbf{OPRL}} & \multirow{2}{*}{\textbf{DPPO}} & \multirow{2}{*}{\textbf{SeqRank}} & \multirow{2}{*}{\textbf{\makecell[r]{LiRE}}}  \\
			& & & &  \\
			\midrule
			\midrule
\multirow{2}{*}{button-press-topdown}  & $\text{exp}(x)$ & 9.60 \tiny{$\pm$ 5.74 } & 22.87 \tiny{$\pm$ 9.06 }& 12.13 \tiny{$\pm$ 5.75 } & 3.93 \tiny{$\pm$ 4.34 } &  20.00 \tiny{$\pm$ 3.54 }& 12.87 \tiny{$\pm$ 7.86 }\\
& $x$ & 36.87 \tiny{$\pm$ 13.75}  & 17.33 \tiny{$\pm$ 10.7 } &  30.00 \tiny{$\pm$ 7.21 } &  1.33 \tiny{$\pm$ 2.31 } & 54.87 \tiny{$\pm$ 9.89 } & 67.20 \tiny{$\pm$ 18.97}	 \\ 
\midrule
\multirow{2}{*}{box-close}  & $\text{exp}(x)$ &  10.33 \tiny{$\pm$ 8.23 }& 0.33 \tiny{$\pm$ 1.16 }&  	4.73 \tiny{$\pm$ 3.24 }& 10.20 \tiny{$\pm$ 11.47} & 26.8 \tiny{$\pm$ 3.47 }& 22.73 \tiny{$\pm$ 10.40} \\
& $x$ &  11.27 \tiny{$\pm$ 14.91} & 1.33 \tiny{$\pm$ 2.31 } &  47.33 \tiny{$\pm$ 39.71} & 3.33 \tiny{$\pm$ 4.16 } & 46.67 \tiny{$\pm$ 36.49} & 51.53 \tiny{$\pm$ 18.48} \\ 
\midrule
\multirow{2}{*}{dial-turn}  & $\text{exp}(x)$ &  50.20 \tiny{$\pm$ 8.51 }   & 68.67 \tiny{$\pm$ 12.39}  & 54.33 \tiny{$\pm$ 11.47} & 26.67 \tiny{$\pm$ 22.23} & 62.27 \tiny{$\pm$ 5.97 }& 65.87 \tiny{$\pm$ 9.46 }\\
& $x$ &  77.27 \tiny{$\pm$ 11.9 }& 56.67 \tiny{$\pm$ 10.87} & 71.33 \tiny{$\pm$ 9.87 } & 22.0 \tiny{$\pm$ 21.07}  &  59.80 \tiny{$\pm$ 17.73} & 79.07 \tiny{$\pm$ 10.96} \\ 
\midrule
\multirow{2}{*}{lever-pull}  & $\text{exp}(x)$ & 50.53 \tiny{$\pm$ 8.55 } & 82.40 \tiny{$\pm$ 22.69} & 96.0 \tiny{$\pm$ 4.00 }& 10.13 \tiny{$\pm$ 12.19}  & 97.07 \tiny{$\pm$ 1.80 } & 57.87 \tiny{$\pm$ 11.28} \\
& $x$ &  70.20 \tiny{$\pm$ 18.03}  & 78.8 \tiny{$\pm$ 13.87} & 86.67 \tiny{$\pm$ 12.2 } &  4.00 \tiny{$\pm$ 5.67 }& 74.33 \tiny{$\pm$ 18.50} & 95.67 \tiny{$\pm$ 6.26 }\\ 
		\bottomrule
		\end{tabular}
	\end{adjustbox}
\vspace{-.1in}
\end{table*}

\subsection{Online PbRL}
\begin{figure*}[ht]
\centering
\subfigure[dial-turn]{
\includegraphics[width=.20\columnwidth]{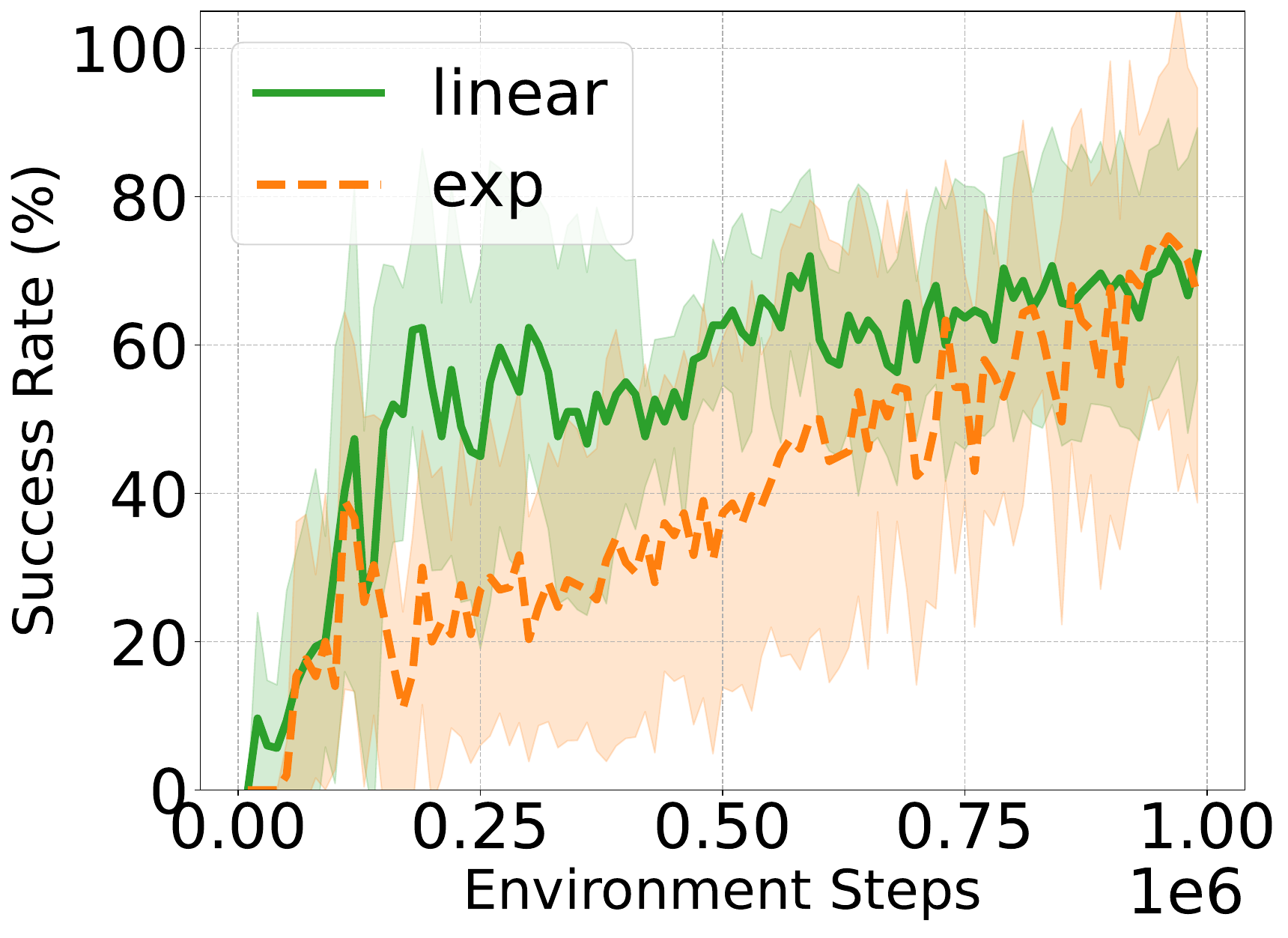}
\label{fig:onlinePbRL1}
}
\subfigure[hand-insert]{
\includegraphics[width=.20\columnwidth]{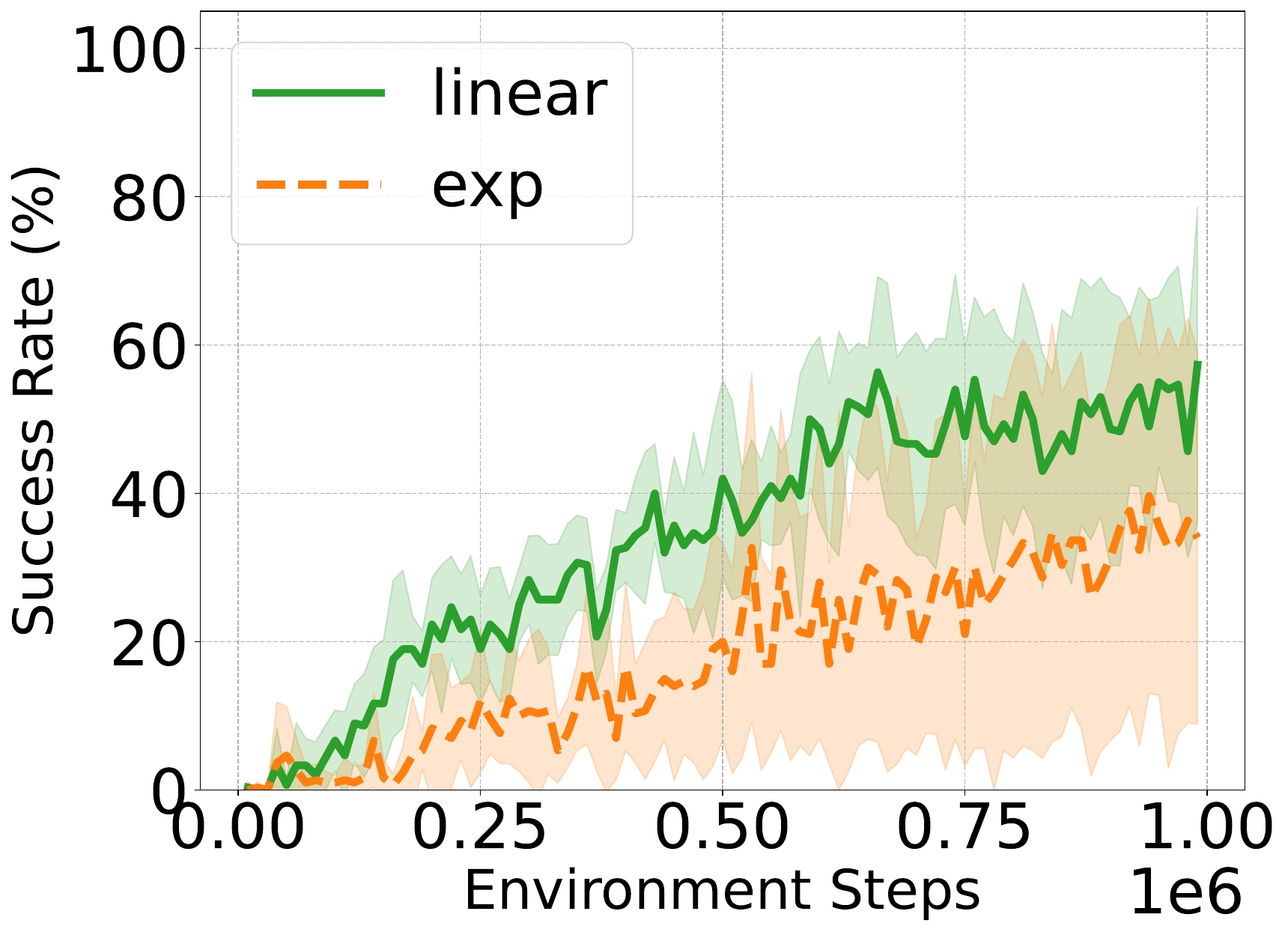}
\label{fig:onlinePbRL2}
}
\subfigure[sweep-into]{
\includegraphics[width=.20\columnwidth]{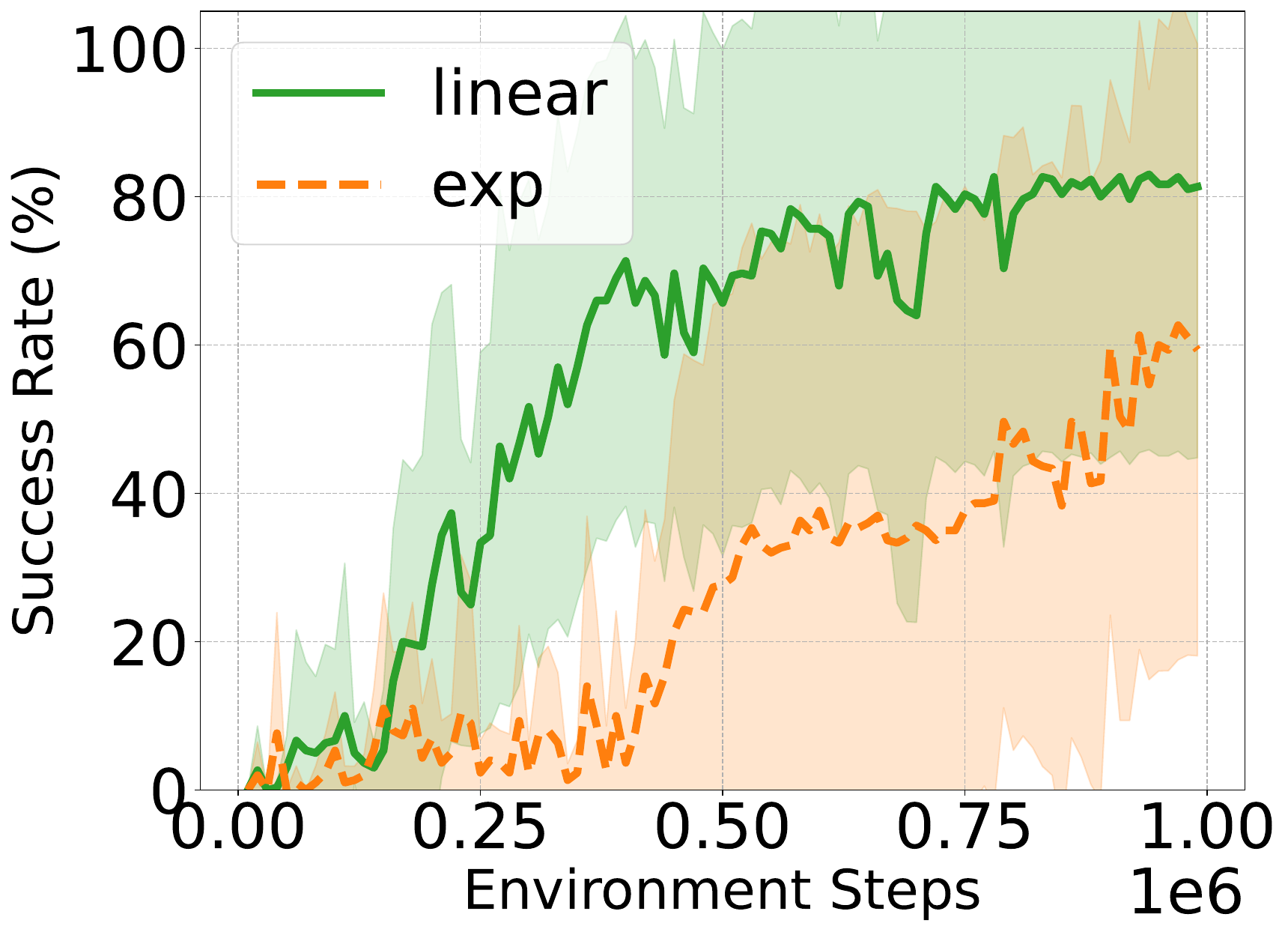}
\label{fig:onlinePbRL3}
}
\subfigure[lever-pull]{
\includegraphics[width=.20\columnwidth]{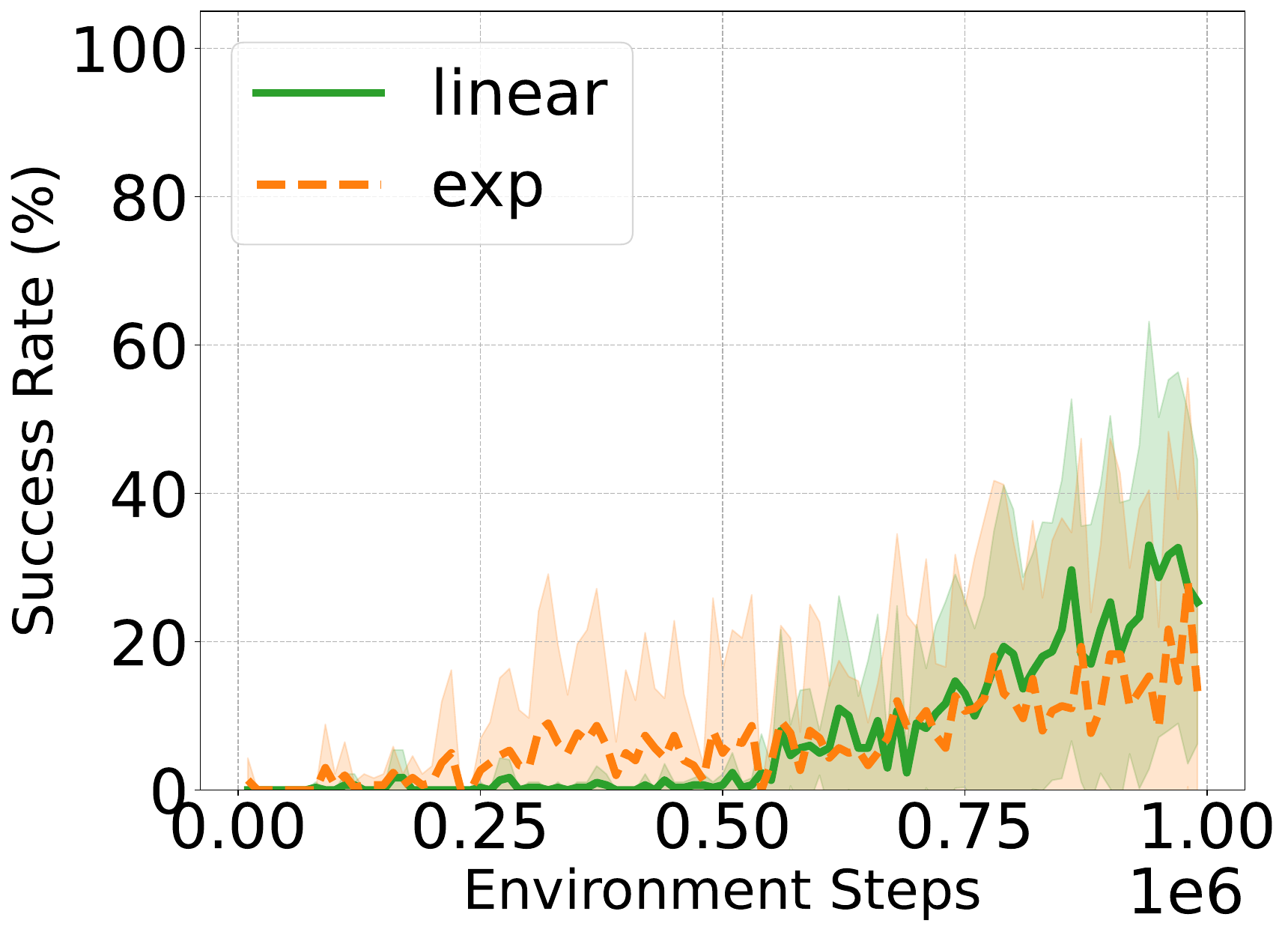}
\label{fig:onlinePbRL}
}
\caption{
Learning curves on online PbRL experiments. Using linear score function in online PbRL improves performance.
}
\label{fig:appendix_online}
\vspace{-.2in}
\end{figure*}

\cref{fig:appendix_online} depicts the experimental results of online PbRL.
We compare the online PbRL performance by using a linear score function and an exponential score function.
The increase in performance when using the linear score function suggests that the BT model using the exponential score function may not be the optimum choice for PbRL. We used the code implemented in PEBBLE \cite{lee2021pebble}.

\section{Details of the Main Experimental Results}
\subsection{Full Learning Curves of Each Method}

\begin{figure*}[!ht]
  \centering
  \includegraphics[width=0.95\textwidth]{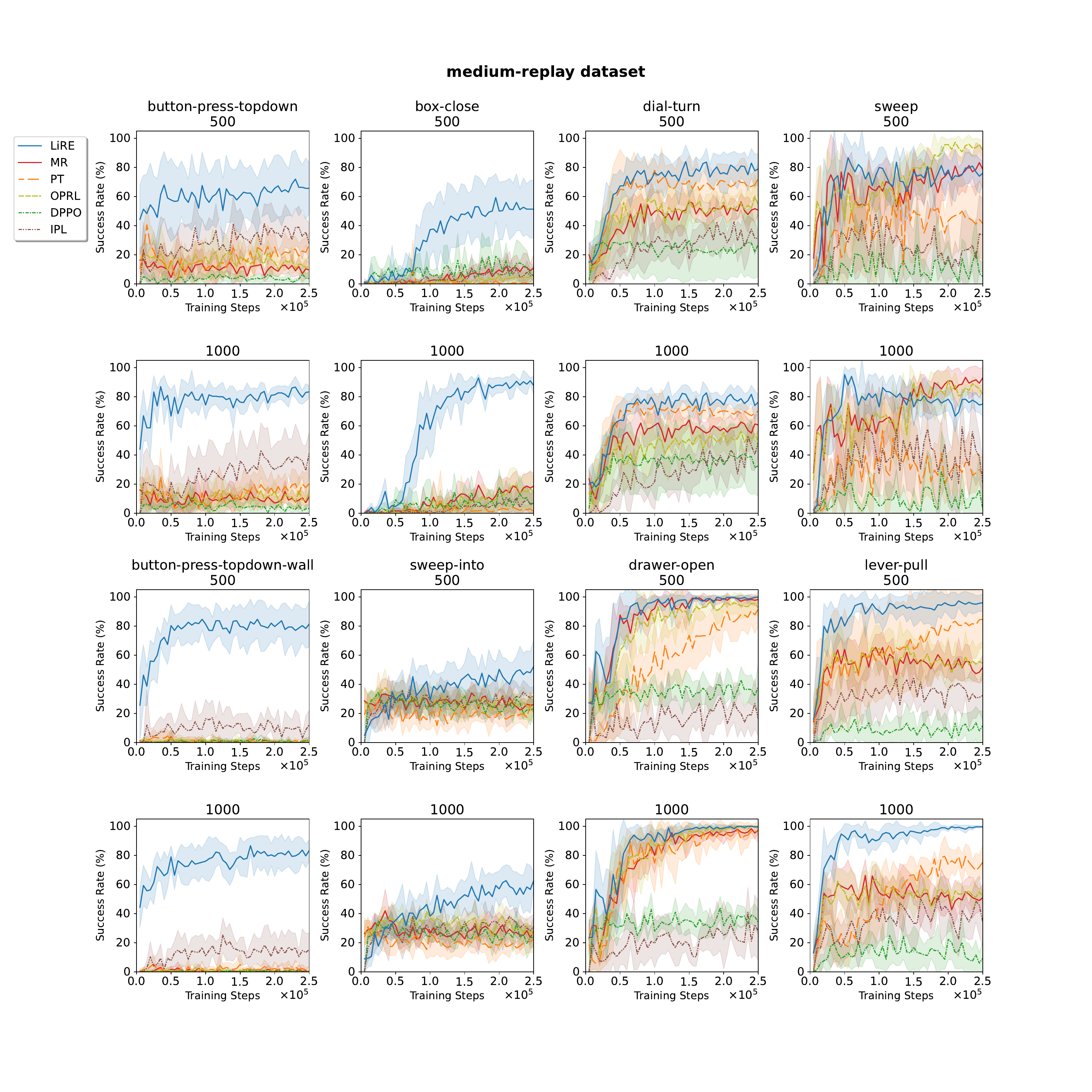}
  \caption{Full learning curves of baselines and LiRE for the Meta-World \texttt{medium-replay} dataset. We use $500$ and $1000$ preference feedbacks and LiRE significantly outperforms existing algorithms for many tasks.}
  \label{fig:full_mr}
  \vspace{-.2in}
\end{figure*}

\begin{figure*}[!ht]
  \centering
  \includegraphics[width=0.65\textwidth]{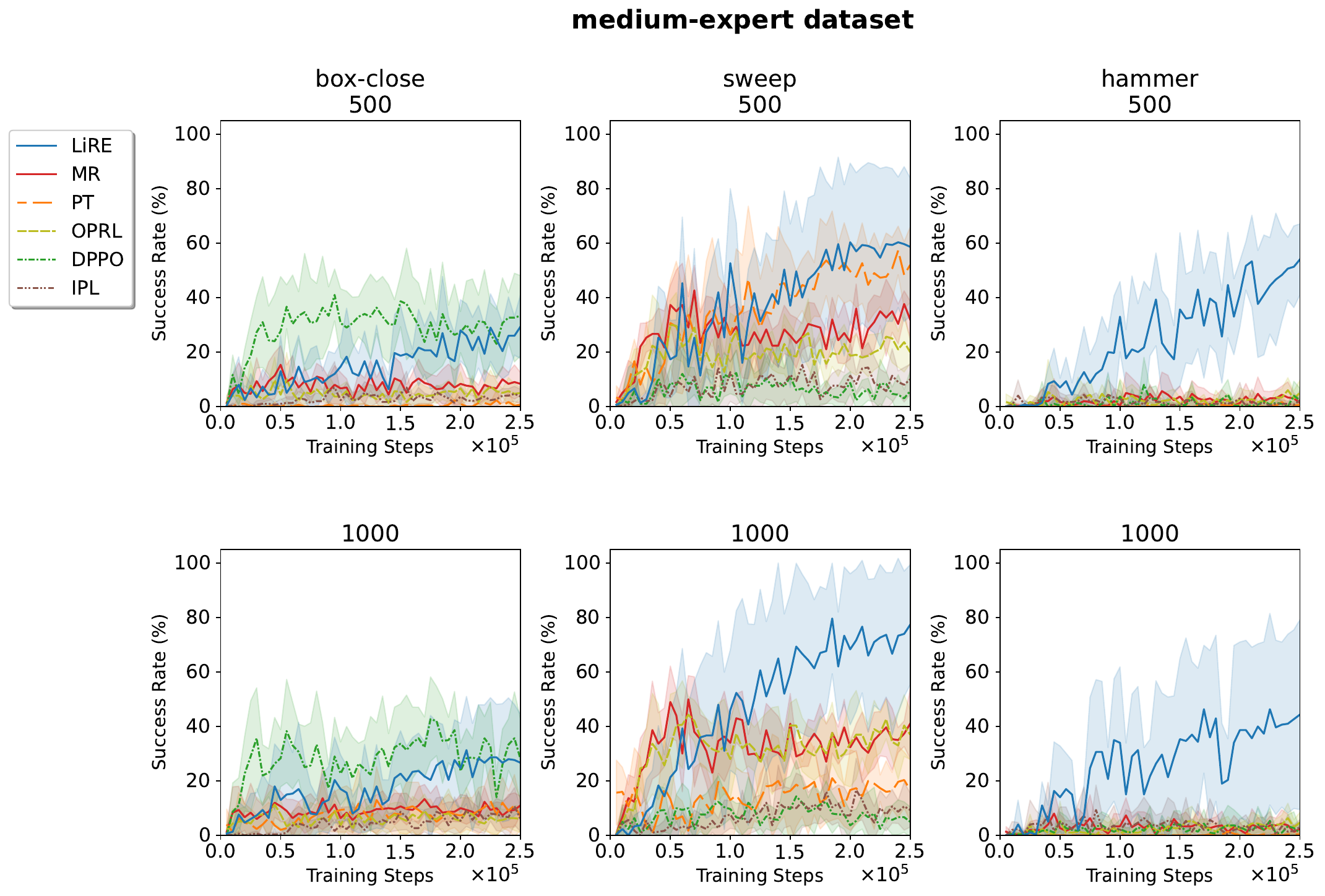}
  \caption{Full learning curves of baselines and LiRE for the Meta-World \texttt{medium-expert} dataset. We use $500$ and $1000$ preference feedbacks. LiRE significantly outperforms existing algorithms for \textit{sweep} and \textit{hammer} tasks.}
  \label{fig:full_me}
  \vspace{-.1in}
\end{figure*}

Full learning curves for the Meta-World \texttt{medium-replay} dataset are shown in \cref{fig:full_mr} and for the Meta-World \texttt{medium-expert} dataset are shown in \cref{fig:full_me}.
We plot the results for MR, PT \cite{kim2022preference}, OPRL \cite{shin2022benchmarks}, DPPO \cite{an2023direct}, IPL \cite{hejna2023inverse}, and LiRE.
The average success rates reported in \cref{tab:baselines} are obtained with the last 5 trained policies.
Although the performance of LiRE for the \texttt{medium-replay} \textit{sweep} task is relatively low, the full learning curve shows that the performance of the best-trained policy is competitive.

\subsection{Ablation Study of LiRE}
We evaluate the success rate by the following: (1) with MR or with LiRE and (2) exponential or linear score function.
\cref{tab:appendix_linear} shows that both constructing RLT and using linear function improve offline PbRL performance. 
\begin{table}[!ht]
	\caption{Average success rates on Meta-World \texttt{medium-replay} dataset with $500$ preference feedbacks. We train the reward model with MR or LiRE using exponential or linear score functions.}
	\label{tab:appendix_linear}
	\vskip 0.15in
	\begin{center}
	\begin{small}
	\resizebox{0.7\columnwidth}{!}{
	\begin{tabular}{l|cccc|}
	\multicolumn{1}{c|}{\multirow{2}{*}{\textbf{Task}}}                         & \multicolumn{2}{c|}{$\phi(x)=\text{exp}(x)$}                       & \multicolumn{2}{c}{\textbf{$\phi(x)=x$}}  \\ \cline{2-5}  
	\multicolumn{1}{c|}{}                         & \multicolumn{1}{c}{MR} & \multicolumn{1}{c|}{LiRE} & \multicolumn{1}{c}{MR} & \multicolumn{1}{c}{LiRE} \\ \cline{2-5} \hline \hline
	button-press-topdown                          & \multicolumn{1}{r}{9.60 \tiny{$\pm$ 5.74}} & \multicolumn{1}{r|}{12.87 \tiny{$\pm$ 7.86}} & \multicolumn{1}{r}{{36.87} \tiny{$\pm$ 13.75}}  & \multicolumn{1}{r}{{67.20} \tiny{$\pm$ 18.97}} \\ 
	box-close                                     & \multicolumn{1}{r}{10.33 \tiny{$\pm$ 8.23}} & \multicolumn{1}{r|}{{22.73} \tiny{$\pm$ 10.40}} & \multicolumn{1}{r}{11.27 \tiny{$\pm$ 14.91}} & \multicolumn{1}{r}{{51.53} \tiny{$\pm$ 18.48}} \\ 
	dial-turn                                     & \multicolumn{1}{r}{50.20 \tiny{$\pm$ 8.51}}  & \multicolumn{1}{r|}{65.87 \tiny{$\pm$ 9.46}} & \multicolumn{1}{r}{{77.27} \tiny{$\pm$ 11.90}}& \multicolumn{1}{r}{{79.07} \tiny{$\pm$ 10.96}} \\ 
	sweep                                    & \multicolumn{1}{r}{{79.80} \tiny{$\pm$ 13.36}}  & \multicolumn{1}{r|}{{82.67} \tiny{$\pm$ 19.86}} & \multicolumn{1}{r}{78.47 \tiny{$\pm$ 10.88}} & \multicolumn{1}{r}{77.53 \tiny{$\pm$ 10.50}} \\
	button-press-topdown-wall                                    & \multicolumn{1}{r}{0.13 \tiny{$\pm$ 0.50}}  & \multicolumn{1}{r|}{1.33 \tiny{$\pm$ 2.15}} & \multicolumn{1}{r}{{8.27} \tiny{$\pm$ 8.64}} & \multicolumn{1}{r}{{79.13} \tiny{$\pm$ 15.19}} \\
	sweep-into                                    & \multicolumn{1}{r}{24.80 \tiny{$\pm$ 5.28}} & \multicolumn{1}{r|}{24.87 \tiny{$\pm$ 8.39}}  & \multicolumn{1}{r}{{49.73} \tiny{$\pm$ 13.52}} & \multicolumn{1}{r}{{49.13} \tiny{$\pm$ 15.85}} \\
	drawer-open                                    & \multicolumn{1}{r}{98.07 \tiny{$\pm$ 3.20}}  & \multicolumn{1}{r|}{{98.67} \tiny{$\pm$ 1.89}} & \multicolumn{1}{r}{97.20 \tiny{$\pm$ 5.88}} & \multicolumn{1}{r}{{99.40} \tiny{$\pm$ 1.65}} \\
	lever-pull                                    & \multicolumn{1}{r}{50.53 \tiny{$\pm$ 8.55}} & \multicolumn{1}{r|}{57.87 \tiny{$\pm$ 11.28}} & \multicolumn{1}{r}{{70.20} \tiny{$\pm$ 18.03}} & \multicolumn{1}{r}{{95.67} \tiny{$\pm$ 6.26}} \\
	\end{tabular}
	 }
	\end{small}
	\end{center}
	\vspace{-.2in}
	\end{table}

\subsection{Effect of RLT and Score Function on Reward Estimation}
\begin{figure*}[!ht]
\centering
\subfigure[MR w/ exp]{
\includegraphics[width=.2\columnwidth]{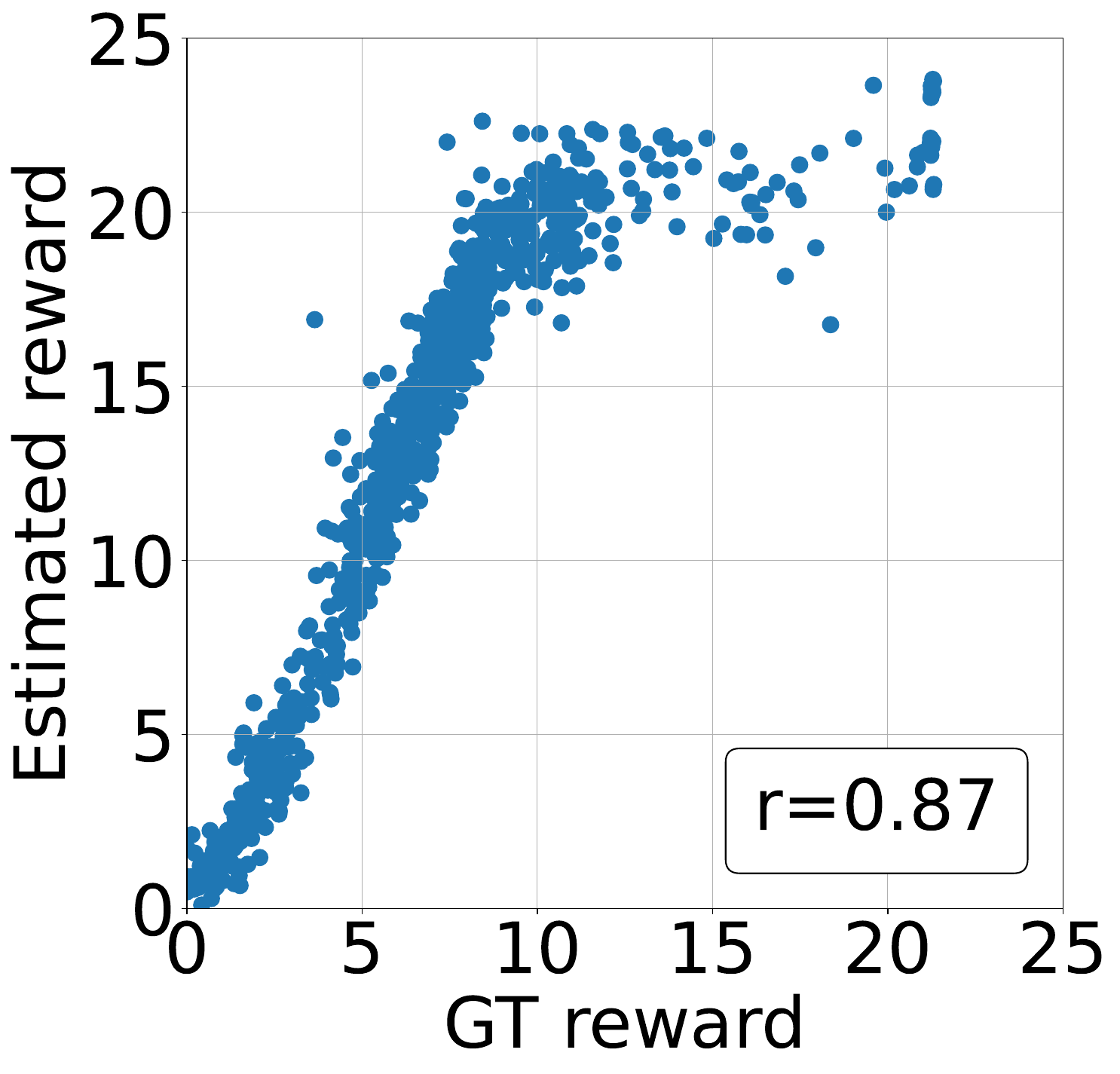}
}
\subfigure[MR w/ linear]{
\includegraphics[width=.2\columnwidth]{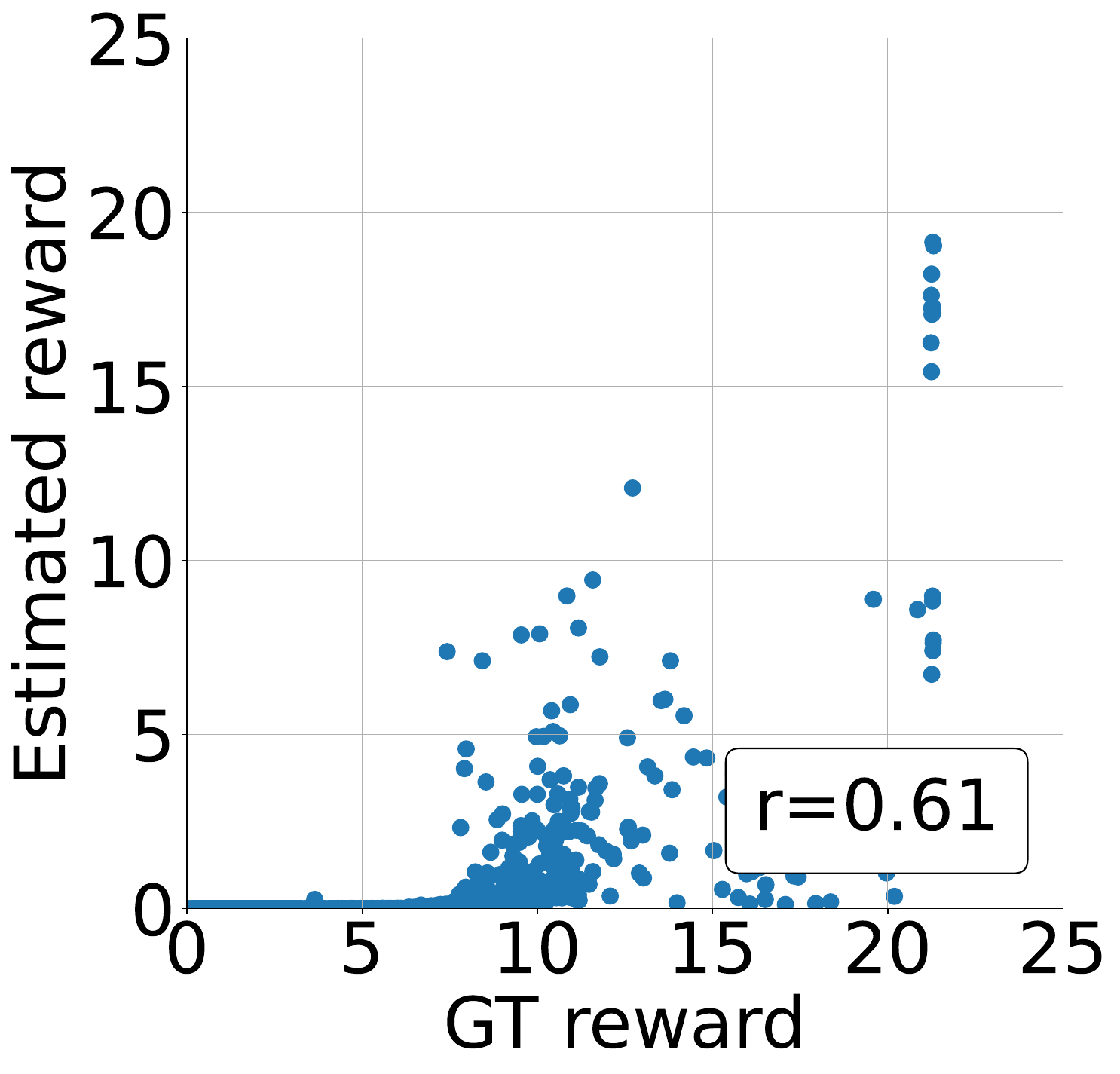}
\label{fig:appendix_reward1}
}
\subfigure[LiRE w/ exp]{
\includegraphics[width=.2\columnwidth]{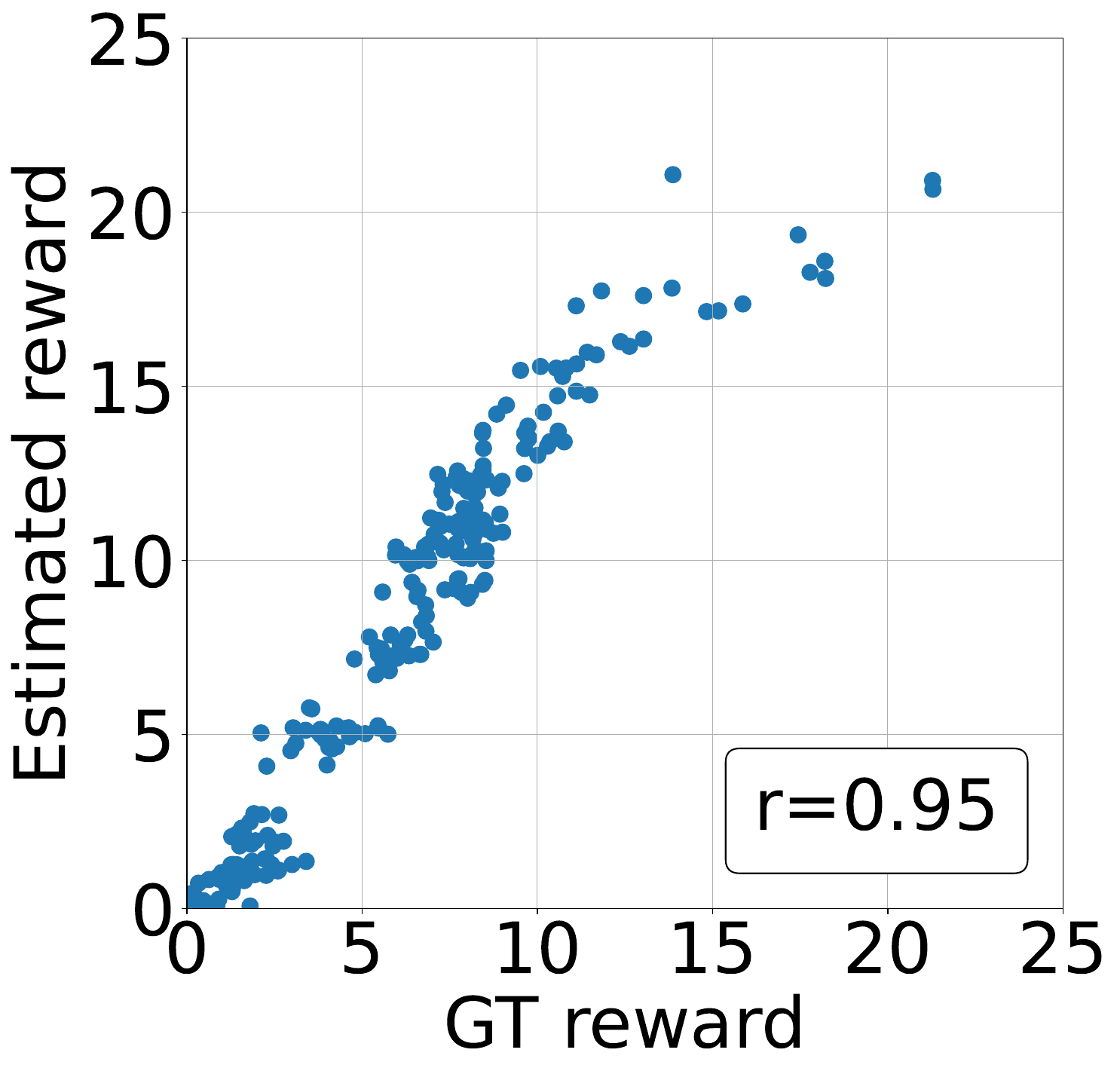}
}
\subfigure[LiRE w/ linear]{
\includegraphics[width=.2\columnwidth]{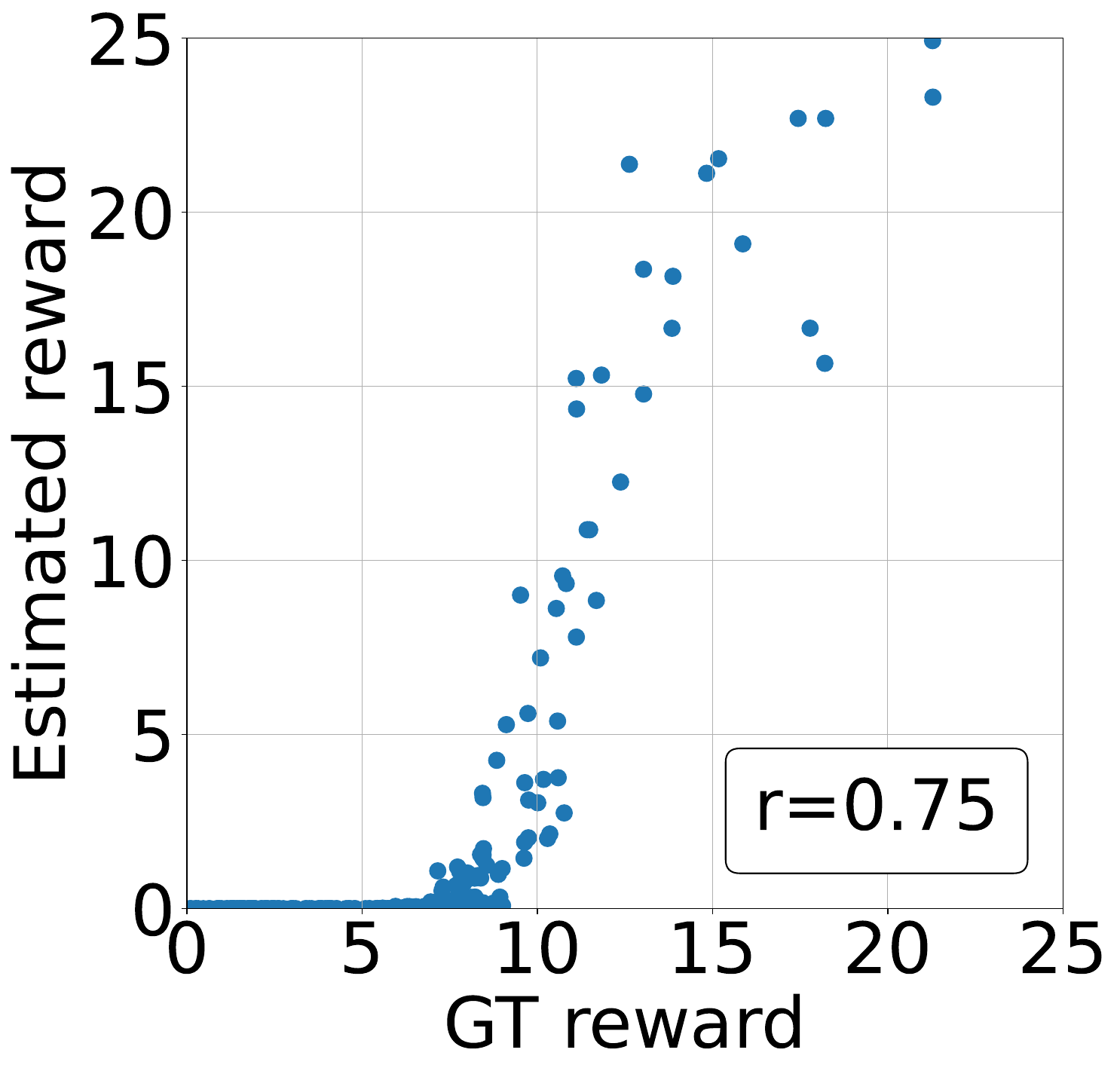}
}
\caption{
Estimated rewards for the segments used in preference learning for \textit{box-close} task. We train the reward model with MR or LiRE using the exp or linear score function. The Pearson correlation coefficient, $r$ is presented.}
\label{fig:appendix_reward}
\vspace{-.1in}
\end{figure*}
Similar to \cref{fig:reward}, in \textit{button-press-topdown} task, \cref{fig:appendix_reward} shows that constructing RLT and using a linear score function can better distinguish the rewards between segments with relatively high preference.

\subsection{Comparison with SeqRank}
\begin{table}[ht]
	\caption{Average success rate and the number of ranks in the list of SeqRank and LiRE.}
	\label{tab:appendix_seq}
	\vskip 0.15in
	\begin{center}
	\begin{small}
	\resizebox{0.7\columnwidth}{!}{
	\begin{tabular}{l|cccc|}
	\multicolumn{1}{c|}{\multirow{3}{*}{\textbf{Task}}} & \multicolumn{4}{c}{\textbf{\makecell{Number of feedacks}}}                         \\ \cline{2-5} 
	\multicolumn{1}{c|}{}                         & \multicolumn{2}{c|}{\textbf{500}}                       & \multicolumn{2}{c}{\textbf{1000}}  \\ \cline{2-5}  
	\multicolumn{1}{c|}{}                         & \multicolumn{1}{c}{SeqRank w/ linear} & \multicolumn{1}{c|}{LiRE} & \multicolumn{1}{c}{SeqRank w/ linear} & \multicolumn{1}{c}{LiRE} \\ \cline{2-5} \hline \hline
	button-press-topdown                          & \multicolumn{1}{c}{54.87 \tiny{$\pm$ 9.89}} & \multicolumn{1}{c|}{67.20 \tiny{$\pm$ 18.97}} & \multicolumn{1}{c}{60.13 \tiny{$\pm$ 11.43}}  & \multicolumn{1}{c}{83.07 \tiny{$\pm$ 6.38}} \\ 
	box-close                                     & \multicolumn{1}{c}{46.67 \tiny{$\pm$ 36.49}} & \multicolumn{1}{c|}{51.53 \tiny{$\pm$ 18.48}} & \multicolumn{1}{c}{65.00 \tiny{$\pm$ 32.90}}  & \multicolumn{1}{c}{89.13 \tiny{$\pm$ 6.02}} \\ 
	dial-turn                                     & \multicolumn{1}{c}{59.80 \tiny{$\pm$ 17.73}} & \multicolumn{1}{c|}{79.07 \tiny{$\pm$ 10.96}} & \multicolumn{1}{c}{45.67 \tiny{$\pm$ 16.47}}  & \multicolumn{1}{c}{76.93 \tiny{$\pm$ 7.55}} \\ 
	sweep                                    	  & \multicolumn{1}{c}{86.33 \tiny{$\pm$ 8.25}} & \multicolumn{1}{c|}{77.53 \tiny{$\pm$ 10.50}} & \multicolumn{1}{c}{94.27 \tiny{$\pm$ 3.92}}  & \multicolumn{1}{c}{75.87 \tiny{$\pm$ 6.81}} \\ 
	button-press-topdown-wall                     & \multicolumn{1}{c}{24.07 \tiny{$\pm$ 16.39}} & \multicolumn{1}{c|}{79.13 \tiny{$\pm$ 15.19}} & \multicolumn{1}{c}{37.07 \tiny{$\pm$ 16.96}}  & \multicolumn{1}{c}{81.47 \tiny{$\pm$ 10.04}} \\ 
	sweep-into                                    & \multicolumn{1}{c}{48.80 \tiny{$\pm$ 18.72}} & \multicolumn{1}{c|}{49.13 \tiny{$\pm$ 15.85}} & \multicolumn{1}{c}{61.53 \tiny{$\pm$ 13.57}}  & \multicolumn{1}{c}{57.73 \tiny{$\pm$ 13.11}} \\ 
	drawer-open                                   & \multicolumn{1}{c}{99.87 \tiny{$\pm$ 0.50}} & \multicolumn{1}{c|}{99.40 \tiny{$\pm$ 1.65}} & \multicolumn{1}{c}{100.00 \tiny{$\pm$ 0.00}}  & \multicolumn{1}{c}{99.73 \tiny{$\pm$ 0.85}} \\ 
	lever-pull                                    & \multicolumn{1}{c}{74.33 \tiny{$\pm$ 18.50}} & \multicolumn{1}{c|}{95.67 \tiny{$\pm$ 6.26}} & \multicolumn{1}{c}{76.27 \tiny{$\pm$ 13.27}}  & \multicolumn{1}{c}{99.47 \tiny{$\pm$ 1.15}} \\ 
	\midrule
	Avg success rate                                       & \multicolumn{1}{c}{61.84 \tiny{$\pm$ 15.80}} & \multicolumn{1}{c|}{\textbf{74.83} \tiny{$\pm$ 12.23}} & \multicolumn{1}{c}{67.49 \tiny{$\pm$ 13.56}}  & \multicolumn{1}{c}{\textbf{82.92} \tiny{$\pm$ 6.48}} \\ 
	Avg \# of ranks in the list                                       & \multicolumn{1}{c}{2.3 \tiny{$\pm$ 0.09}} & \multicolumn{1}{c|}{\textbf{9.3} \tiny{$\pm$ 1.83}} & \multicolumn{1}{c}{2.3 \tiny{$\pm$ 0.09}}  & \multicolumn{1}{c}{\textbf{9.3} \tiny{$\pm$ 1.84}} \\ 
	\end{tabular}
	}
	\end{small}
	\end{center}
	\vspace{-.2in}
	\end{table}
	
\cref{tab:appendix_seq} shows the success rate of each task in \cref{tab:seq}.
SeqRank \cite{hwang2023sequential} improves feedback efficiency but constructs a shorter length of the ranked list, so LiRE is better at utilizing second-order preference.

\section{Experimental Details}
\label{appendix}
\subsection{RLT Construction}
\begin{algorithm}[tb]
   \caption{RLT Construction}
   \label{alg:pref-list}
\begin{algorithmic}
   \FUNCTION{\textsc{BinarySearch} $(\sigma, \texttt{low}, \texttt{high}, L):$}
   \IF{\texttt{low} = \texttt{high}}
        \STATE insert a new group $\{\sigma\}$ to $L$ right behind to $g_{\texttt{low}+1}$
        \STATE (i.e., $g_{\texttt{low}} \prec \{\sigma\} \prec g_{\texttt{low}+1}$)
   \ELSE
        \STATE \textcolor{red}{/* Human Feedback */}
        \STATE compare $\sigma$ to $\sigma_s \in g_{\texttt{mid}}$ where $\texttt{mid} = \big \lceil \frac{\texttt{low}+\texttt{high}}{2} \big \rceil$
        \IF{$\sigma_s \prec \sigma$}
        \STATE \textsc{BinarySearch}($\sigma, \texttt{mid}, \texttt{high}, L$)
        \ELSIF{$\sigma \prec \sigma_s$}
        \STATE \textsc{BinarySearch}($\sigma, \texttt{low}, \texttt{mid}-1, L$)
        \ELSE
        \STATE $g_{\texttt{mid}} \leftarrow g_{\texttt{mid}} \cup \{\sigma\}$
        \ENDIF
   \ENDIF
   \ENDFUNCTION
   \medskip
   \STATE {\bfseries Init:} List $L=[\,]$
   \REPEAT
   \STATE sample $\sigma_1, \sigma_2, \cdots \in D_s$
   \IF{$L$ is empty}
   \STATE $L \leftarrow [\{\sigma_i\}]$
   \ELSE
   \STATE \textsc{BinarySearch}$(\sigma_i, 0, l, L)$
   \ENDIF
   \UNTIL{end of feedback}
   \STATE {\bfseries Output:} $L$
\end{algorithmic}
\end{algorithm}

    
            
            


    

To construct RLT, We can use any sorting method such as binary insertion sort, mergesort, or quicksort.
However, if the RLT is already partially constructed, a binary insertion sort is an efficient way to find the rank of each segment. 
The pseudocode for the binary insertion sort we use to construct the RLT is summarized in \cref{alg:pref-list}.

\subsection{Creating Offline PbRL Dataset}
\label{appendix-dataset}
Following offline RL data collection approach, we collect offline RL data from different policies in two ways: \texttt{medium-replay} dataset and \texttt{medium-expert} dataset.

\textbf{\texttt{medium-replay} dataset}
We use the replay buffer collected while training online RL as an offline RL dataset.
We train with 3 seeds using the online SAC \cite{haarnoja2018soft} implemented in PEBBLE \cite{lee2021pebble} with ground-truth rewards.
We stop collecting replay buffers when the average success rate of the online RL's performance is near 50.
(For the DMControl dataset, collect until the episode returns are about in the middle of the convergence value.)
We measure the online RL performance every 50,000 steps, so depending on the training speed of the online RL, the average success rate of the online RL may be less or more than 50 at the end of the replay buffer collection.
\cref{table: medium-replay dataset} shows the average success rate when the collection of the replay buffers ends.

\begin{table*}[ht]
  \small
  \caption{Average success rate of online RL when replay buffer collection ends.}
  \centering
  \begin{tabular}{cccc}
    \specialrule{0.12em}{0pt}{0pt}
	button-press-topdown 
	& button-press-topdown-wall 
	& box-close
	& dial-turn
	\\ \hline 
    47.0  & 36.0  & 46.0    & 49.3 
    \\ \hline
	sweep-v2
	& sweep-into
	& drawer-open
	& lever-pull
	\\ \hline
    32.0  &  55.3 & 61.3  & 78.0  \\
    \specialrule{0.12em}{0pt}{0pt}
  \end{tabular}
  \label{table: medium-replay dataset}
\end{table*}

\textbf{\texttt{medium-expert} dataset}
We collect \texttt{medium-expert} dataset following approaches by prior works \cite{hejna2023inverse, anonymous2023efficient}:
collect 50 trajectories from the expert policy provided by Meta-World \cite{yu2020meta}, collect 50 trajectories from the expert policy for a different randomized object and goals positions, 
collect 100 trajectories from the expert policy for a different task out of 50 Meta-World tasks,
collect 200 trajectories from a random policy, and finally, collect 200 trajectories from the $\epsilon$-greedy policy that samples an action from the expert policy with 50\% probability and from the random policy with the remaining 50\% probability.
We also add Gaussian noise with a mean of 0 and a standard deviation of 1 for each policy.

\subsection{RL Performance between GT Reward and Wrong Rewards}
For each dataset, we verify that there is a difference in RL performance when trained with GT reward versus wrong rewards because if offline RL achieves high performance with wrong rewards, the dataset is not appropriate for offline PbRL.
We use the three wrong rewards chosen by \cite{li2023survival}: zero rewards, where all rewards $r(s,a)=0$; random rewards, where all reward values are sampled from a uniform distribution $U(0,1)$; and negative rewards, set to $-r(s,a)$. 
The performance of offline RL with GT reward and wrong rewards on each dataset is shown in \cref{table: GT and wrong} and \cref{table: DMC GT and wrong}.

\begin{table}[!ht]
	\vspace{-1pt}
	\caption{Average success rate of each dataset on GT rewards and wrong rewards with IQL \cite{kostrikov2021offline}.}
	\centering
	\resizebox{0.8\columnwidth}{!}{
	\begin{tabular}{clrrrr}
	\toprule
	& Task & GT & Zero & Random & Negative \\
	\midrule
	\multirow{8}{*}{medium-replay dataset}        & button-press-topdown & 88.33 \tiny{$\pm$ 4.76 }& 12.07 \tiny{$\pm$ 5.76 }& 13.00 \tiny{$\pm$ 5.36 }& 0.00 \tiny{$\pm$ 0.00 }\\
										  & box-close & 93.40 \tiny{$\pm$ 3.10 }& 0.53 \tiny{$\pm$ 0.88 }& 0.13 \tiny{$\pm$ 0.50 }& 0.13 \tiny{$\pm$ 0.50 }\\
										  & dial-turn  & 75.40 \tiny{$\pm$ 5.47 }& 16.07 \tiny{$\pm$ 6.44 }& 13.93 \tiny{$\pm$ 7.70 }& 2.40 \tiny{$\pm$ 3.32}\\
										  & sweep  & 98.33 \tiny{$\pm$ 1.87 }& 0.20 \tiny{$\pm$ 0.60 }& 0.40 \tiny{$\pm$ 0.80 }& 0.00 \tiny{$\pm$ 0.00}\\
										  & button-press-topdown-wall & 56.27 \tiny{$\pm$ 6.32 }& 1.67 \tiny{$\pm$ 1.64 }& 1.13 \tiny{$\pm$ 1.77 }& 0.00 \tiny{$\pm$ 0.00 }\\
										  & sweep-into & 78.80 \tiny{$\pm$ 7.96 }& 24.73 \tiny{$\pm$ 7.26 }& 23.40 \tiny{$\pm$ 7.23 }& 0.07 \tiny{$\pm$ 0.36 }\\
										  & drawer-open  & 100.00 \tiny{$\pm$ 0.00 }& 25.67 \tiny{$\pm$ 10.65} & 22.33 \tiny{$\pm$ 11.66} & 0.00 \tiny{$\pm$ 0.00}\\
										  & lever-pull  & 98.47 \tiny{$\pm$ 1.77 }& 1.27 \tiny{$\pm$ 1.50} & 1.20 \tiny{$\pm$ 1.51} & 0.00 \tiny{$\pm$ 0.00} \\
	\midrule
	\multirow{3}{*}{medium-expert dataset} & box-close & 65.00 \tiny{$\pm$ 9.98 }& 3.67 \tiny{$\pm$ 4.68 }& 2.67 \tiny{$\pm$ 3.77 }& 1.00 \tiny{$\pm$ 1.53 }\\
										  & sweep  & 85.33 \tiny{$\pm$ 5.96 }& 5.00 \tiny{$\pm$ 10.31} & 2.00 \tiny{$\pm$ 3.65 } & 0.00 \tiny{$\pm$ 0.00 }\\
										  & hammer  & 65.00 \tiny{$\pm$ 11.36} & 2.33 \tiny{$\pm$ 5.22 }& 1.67 \tiny{$\pm$ 3.73 }& 1.33 \tiny{$\pm$ 2.21}\\
	\bottomrule
	\end{tabular}
	}
	\vspace{-.1in}
	\label{table: GT and wrong}
	\end{table}

\begin{table}[!ht]
\vspace{-1pt}
\caption{Episode returns of each DMControl \texttt{medium-replay} dataset on GT rewards and wrong rewards with IQL \cite{kostrikov2021offline}.}
\centering
\resizebox{0.6\columnwidth}{!}{
\begin{tabular}{lrrrr}
\toprule
Task & GT & Zero & Random & Negative \\
\midrule
hopper-hop & 157.95 \tiny{$\pm$ 9.64 }& 18.9 \tiny{$\pm$ 7.5} & 19.79 \tiny{$\pm$ 7.47 }&  0.01 \tiny{$\pm$ 0.02 }\\
walker-walk & 839.6 \tiny{$\pm$ 36.57} & 189.58 \tiny{$\pm$ 28.15} & 234.14 \tiny{$\pm$ 37.22} & 28.79 \tiny{$\pm$ 2.2 }\\
humanoid-walk  & 250.9 \tiny{$\pm$ 11.62} & 60.36 \tiny{$\pm$ 10.56} & 65.13 \tiny{$\pm$ 10.16} & 1.38 \tiny{$\pm$ 0.21 }\\
\bottomrule
\end{tabular}
}
\vspace{-.2in}
\label{table: DMC GT and wrong}
\end{table}

\subsection{Preference Label}
\label{appendix-details}
We set the length of segment $\sigma$ used in the preference label to 25, denoted as $T=25$ in $\sigma=(s_0, a_1, \cdots, s_{T-1}, a_{T-1})$.
We use the GT reward to label the preference between segment pairs.
Considering that GT reward in Meta-World ranges from 0 to 10, segments with GT reward differences less than $12.5$ are labeled as equally preferred segments.
This threshold is equivalent to the threshold provided by B-pref \cite{lee2021b}, which is used as an online PbRL benchmark, when the policy has an average return of $5$ (that is, medium performance).

\subsection{Hyperparameters}
\label{appendix:hyperparameters}
\textbf{Reward model}
The reward model used in our method and the standard pairwise PbRL and MR reward model use the same reward model structure.
We ensemble three reward models and finally predicted the reward in the offline RL dataset by averaging the estimated reward values from the three reward models.
The details of the hyperparameters are shown in \cref{table: hyperparameter}.

In our experiments, MR, PT, and OPRL are two-step PbRL methods that first train the reward model and learn the offline RL with the trained reward model.
We use the trained reward model to estimate the reward for every $(s,a)$ in the offline RL dataset and apply min-max normalization to the reward values in the dataset so that the minimum and maximum values are 0 and 1.
We also apply min-max normalization to the experiments with GT rewards and wrong rewards for a fair comparison.

\textbf{Implementation details}
We choose IQL for the default offline RL algorithm and CORL \cite{tarasov2022corl} for the implementation code\footnote{\href{https://github.com/tinkoff-ai/CORL}{https://github.com/tinkoff-ai/CORL}}.
We use IQL because it is the default offline RL algorithm in previous offline PbRL papers, and IQL is also one of the strongest offline algorithms according to CORL.
We use the same hyperparameters that were used to train Gym-MuJoCO in CORL.
For PT, we follow their implementation\footnote{\href{https://github.com/csmile-1006/PreferenceTransformer}{https://github.com/csmile-1006/PreferenceTransformer}} for the training reward model and use the CORL library for training offline RL.
We follow the official implementations of DPPO\footnote{\href{https://github.com/snu-mllab/DPPO}{https://github.com/snu-mllab/DPPO}} and IPL\footnote{\href{https://github.com/jhejna/inverse-preference-learning}{https://github.com/jhejna/inverse-preference-learning}} with the hyperparameters they use in the Gym-MuJoCo and Metaworld dataset.
The hyperparameters for each baseline, including IQL, are listed in \cref{table: hyperparameter}.
The total number of gradient descent steps in the offline RL is 250,000 and we evaluate the success rate for 50 episodes every 5000 steps.
We run six seeds for all baselines and our method.
We then report the average success rate of the last 5 trained policies.
We use a single NVIDIA RTX A5000 GPU and 32 CPU cores (AMD EPYC 7513 @ 2.60GHz) in our experiments.

\begin{table}[ht]
\vspace{-1pt}
\caption{Hyperparameters of the reward model and the baselines.}
\centering
\begin{tabular}{cll}
\toprule
& Hyperparameter & Value \\
\midrule
\multirow{11}{*}{Reward model}        & Optimizer & Adam \cite{kingma2014adam} \\
                                      & Learning rate & 1e-3 \\
                                      & Batch size  & 512 \\
                                      & $Q$ & 100 \\
                                      & Hidden layer dim      & 128 \\
                                      & Hidden layers   & 3 \\
                                      & Activation function    & ReLU \\
                                      & Final activation & Tanh \\
                                      & Epochs & 300 \\
                                      & \# of ensembles & 3 \\
                                      & Reward from the ensemble models & Average \\
\midrule
\multirow{9}{*}{IQL \cite{kostrikov2021offline}}     & Optimizer & Adam \cite{kingma2014adam} \\
                                      & Critic, Actor, Value hidden dim    & 256        \\
                                      & Critic, Actor, Value hidden layers & 2          \\
                                      & Critic, Actor, Value activation function & ReLU \\
                                      & Critic, Actor, Value learning rate & 0.5 \\
                                      & Mini-batch size & 256 \\
                                      & Discount factor & 0.99 \\
                                      & $\beta$ & 3.0 \\
                                      & $\tau$ & 0.7 \\
\midrule
\multirow{4}{*}{PT \cite{kim2022preference}}    & Optimizer    & AdamW \cite{loshchilov2018decoupled} \\                
                                         & \# of layers    & 1 \\
                                         & \# of attention heads            & 4 \\
                                         & Embedding dimension            & 256 \\
                                         & Dropout rate            & 0.1 \\
                                         
\midrule
\multirow{6}{*}{IPL \cite{hejna2023inverse}}  & Optimizer & Adam \cite{kingma2014adam} \\
                                      & Regularization $\lambda$ & 3e-4 \\
                                      & $Q, V, \pi$ arch      & 3x256d \\
                                      & $\beta$         & 4.0 \\
                                      & $\tau$        & 0.7 \\
                                      & Subsample $s$ & 16 \\
\midrule
\multirow{4}{*}{DPPO \cite{an2023direct}}    
                                      &  Preference predictor   & The same as PT \cite{kim2022preference}     \\
                                      &  Smoothness regularization $\nu$   & 1.0      \\
                                      &  Smoothness sigma $m$  & 20       \\
                                      & Regularization $\lambda$  & 0.5       \\
\midrule
\multirow{4}{*}{OPRL \cite{shin2022benchmarks}}            
                                      & \# of ensembles          & 7    \\
                                      & Initial preference labels       & 30\% of feedback budget     \\
                                      & Every 50 epochs        & 10\% of feedback budget \\  
                                      & Total epochs & 500 \\
\bottomrule
\end{tabular}
\label{table: hyperparameter}
\end{table}


\end{document}